\documentclass[3p, twocolumn]{elsarticle}

\usepackage{amsmath, graphicx, booktabs, tabularx, anyfontsize, framed}
\usepackage{xcolor}
\usepackage[colorlinks = true,
            linkcolor = blue,
            urlcolor  = blue,
            citecolor = blue,
            anchorcolor = blue]{hyperref}

\usepackage[modulo]{lineno}
\modulolinenumbers[2]

\graphicspath{{./figures/}}

\journal{Medical Image Analysis}

\biboptions{authoryear}

\begin{document}

\begin{frontmatter}

\title{Building medical image classifiers with very limited data using segmentation networks\tnoteref{t1}}
\tnotetext[t1]{This paper was accepted by Medical Image Analysis. The publication is available via \url{https://doi.org/10.1016/j.media.2018.07.010}.}


\author{Ken C. L. Wong}
\ead{clwong@us.ibm.com}
\author{Tanveer Syeda-Mahmood}
\ead{stf@us.ibm.com}
\author{Mehdi~Moradi\corref{cor1}}
\ead{mmoradi@us.ibm.com}
\cortext[cor1]{Corresponding author.}

\address{IBM Research -- Almaden Research Center, San Jose, CA, USA}

\begin{abstract}
Deep learning has shown promising results in medical image analysis, however, the lack of very large annotated datasets confines its full potential. Although transfer learning with ImageNet pre-trained classification models can alleviate the problem, constrained image sizes and model complexities can lead to unnecessary increase in computational cost and decrease in performance. As many common morphological features are usually shared by different classification tasks of an organ, it is greatly beneficial if we can extract such features to improve classification with limited samples. Therefore, inspired by the idea of curriculum learning, we propose a strategy for building medical image classifiers using features from segmentation networks. By using a segmentation network pre-trained on similar data as the classification task, the machine can first learn the simpler shape and structural concepts before tackling the actual classification problem which usually involves more complicated concepts. Using our proposed framework on a 3D three-class brain tumor type classification problem, we achieved 82\% accuracy on 191 testing samples with 91 training samples. When applying to a 2D nine-class cardiac semantic level classification problem, we achieved 86\% accuracy on 263 testing samples with 108 training samples. Comparisons with ImageNet pre-trained classifiers and classifiers trained from scratch are presented.
\end{abstract}

\begin{keyword}
image segmentation \sep image classification \sep deep learning \sep transfer learning \sep convolutional neural network \sep fully convolutional network
\end{keyword}

\end{frontmatter}

\section{Introduction}

With the availability of big data and GPU computing, deep learning has become a powerful technique which raises the benchmarks of classification and detection challenges in computer vision \citep{Journal:Schmidhuber:NN2015}. Using huge databases such as ImageNet which has more than a million annotated images \citep{Conference:Deng:CVPR2009}, deep convolutional neural networks (CNNs) such as AlexNet \citep{Conference:Krizhevsky:ANIPS2012}, VGGNet \citep{Journal:Simonyan:arXiv2014}, and GoogLeNet \citep{Conference:Szegedy:CVPR2015} were proposed with impressive performance in visual pattern recognition. In medical imaging, however, such large datasets are usually unavailable especially in 3D analysis. Given the large variety of possible clinical conditions in an imaging modality, it is very challenging to build a sufficiently large dataset with samples of desired abnormalities, and the datasets are usually highly unbalanced. If such datasets are directly used to train classifiers, low classification performance can be expected. Furthermore, different from the annotation tasks of ImageNet, annotations of medical images require radiologists or trained experts to ensure the clinical relevance. Therefore, medical image annotations can be time consuming and expensive. As a consequence, it is highly beneficial if we can reduce the number of annotated samples without decreasing the performance of the classification task.

\subsection{Related work}

To address the above difficulties, transfer learning using ImageNet pre-trained CNNs has been a common approach in medical image analysis \citep{Journal:Litjens:MedIA2017}. In \cite{Conference:van:ISBI2015}, the 4096 off-the-shelf features from the first fully-connected layer of a pre-trained OverFeat model were extracted for pulmonary nodule detection from computed tomography (CT) images through a linear support vector machine (SVM). In \cite{Conference:Moradi:ISBI2016}, hand-crafted features such as the histogram of oriented gradients were combined with features from selected convolutional and fully-connected layers of a pre-trained CNN-M model for cardiac semantic level classification from computed tomography angiography (CTA) images using a SVM. In \cite{Journal:Tajbakhsh:TMI2016}, four different medical imaging applications on three imaging modalities were used to study the differences between CNNs trained from scratch and CNNs fine-tuned from pre-trained models. Using the AlexNet architecture, this study showed that the fine-tuned CNNs performed better as the number of training samples reduced. In \cite{Journal:Shin:TMI2016}, by studying different CNN architectures such as AlexNet, GoogLeNet, and VGGNet on thoraco-abdominal lymph node detection and interstitial lung disease classification, it was shown that models trained from scratch or fine-tuned from ImageNet pre-trained CNNs consistently outperform applications using off-the-shelf CNN features.

Recently, algorithms for building medical image applications on limited training data without ImageNet pre-trained CNNs have been studied. In \cite{Conference:Roy:MICCAI2017}, auxiliary labels generated by an existing automated software tool were used to pre-train a segmentation network. The pre-trained network was fine tuned by error corrective boosting which selectively focuses on classes with erroneous segmentations. In \cite{Conference:Uzunova:MICCAI2017}, by learning a statistical appearance model from few sample images, a large number of synthetic image pairs were generated with the associated ground-truth deformation. These synthetic data were used to fine-tune a pre-trained FlowNet for dense image registration, and it was shown that data-driven, model-based augmentation approach outperforms generic but highly unspecific methods. In \cite{Conference:Madani:ISBI2018}, a semi-supervised learning architecture based on the generative adversarial nets was proposed for cardiac abnormality classification in chest X-rays. This architecture allows both labeled and unlabeled data to contribute to the model thus the required number of labeled data can be largely reduced.

There are several limitations of using ImageNet pre-trained CNNs on medical image analysis. First of all, if the features of the fully-connected layers are used, the input images need to be resized to match the training images (e.g. 224$\times$224$\times$3 for VGGNet). This leads to unnecessary increase in computation for smaller images, or reduced details for larger images. Secondly, the size of the pre-trained model may be unnecessarily large for medical image applications. Using VGGNet as an example, its architecture was proposed to classify 1000 classes of non-medical images. Such a large number of classes is uncommon in medical image analysis and thus such a large model may be unnecessary. Thirdly, to the best of our knowledge, there are no publicly available models equivalent to ImageNet pre-trained CNNs for 3D image analysis. Datasets with millions of annotated 3D medical images are publicly unavailable, and the computational cost of training or using an equivalent model in 3D can be computationally difficult.

\subsection{Our framework}

To overcome the above limitations, here we propose to use a pre-trained segmentation network as the feature source to reduce the number of samples required to train a classification network. By using a segmentation network pre-trained on data similar to those of the classification task, we can achieve high classification performance with very limited training data. The power of this framework can be partly explained by curriculum learning \citep{Conference:Bengio:ICML2009}. Similar to the learning of humans and animals, machine learning can benefit from presenting concepts which are meaningfully organized from simple to complex. In our framework, instead of a classification task which involves complex and abstract concepts such as disease categories, we first train the machine to perform a segmentation task which involves simpler concepts such as shapes and structures. This is similar to radiologists who need to learn anatomical and functional concepts before being able to perform diagnosis. In fact, when pixel-level semantic labels are provided by radiologists or automated software tools for training, the segmentation network learns the anatomical knowledge from experts.

There are several reasons to use a segmentation network as the feature source. First of all, different 2D and 3D segmentation networks are available \citep{Conference:Ronneberger:MICCAI2015,Conference:Milletari:3DV2016,Conference:Mehta:ISBI2017,Journal:Dou:MedIA2017}, and all of them show impressive capabilities of segmenting complicated biological or medical structures with relatively few training samples and straightforward training strategy. In \cite{Conference:Zhou:CVPR2016}, it is shown that classification networks trained on image-level labels can provide useful localization information. For segmentation networks, there are tens of thousands of pixel-level labels providing the shape, structural, and semantic information per image. Therefore, the amount of useful features in a segmentation network can be large even with relatively few training samples.

Secondly, compared with generative models such as the generative adversarial nets \citep{Conference:Goodfellow:NIPS2014,Journal:Mirza:arXiv2014,Conference:Pathak:CVPR2016,Conference:Madani:SPIE2018,Conference:Madani:ISBI2018}, the trainings of segmentation networks are more straightforward and require less data. Furthermore, using features from pre-trained segmentation networks is as simple as using those from pre-trained classification networks. Although training segmentation networks requires pixel-level labels, as our concentration is not on segmentation accuracy, semantic labels generated by existing software tools or even non-semantic labels generated by techniques such as intensity thresholding can provide useful features depending on the complexities of the classification problems.

Using our framework on a three-class brain tumor classification problem of 3D magnetic resonance (MR) images, we achieved 82\% accuracy on 191 testing samples with 91 training samples. When applying to a nine-class semantic level classification problem of 2D cardiac CTA images, we achieved 86\% accuracy on 263 testing samples with 108 training samples.

The preliminary results of this work were reported in \cite{Conference:Wong:MICCAI2017}. Further improvements and experiments are reported in this paper:
\begin{itemize}
  \item The segmentation network in \cite{Conference:Mehta:ISBI2017} is utilized which provides more features than the segmentation network in \cite{Conference:Ronneberger:MICCAI2015} used in \cite{Conference:Wong:MICCAI2017}. A new classification framework is proposed to utilize the features with reduced model size and complexity. Image augmentations are also introduced.

  \item Experiments were performed on 3D brain MR images and 2D cardiac CTA images, not only on disease detection but also on cardiac semantic level classification. This shows that our framework can be applied to different image dimensions, modalities, and classification tasks. The classification problems and reported data in \cite{Conference:Wong:MICCAI2017} were entirely different.

  \item For the cardiac CTA images, instead of training the segmentation network using 2D slices at a particular viewpoint (e.g. the four chamber view), the network is trained on all axial slices of the 3D images. This provides more useful features for complicated tasks such as the cardiac semantic level classification.

  \item Apart from a classification network trained from scratch, we also compare our framework to another network which uses features from ImageNet pre-trained VGGNet. This compares our framework with one of the commonly used transfer learning techniques.
\end{itemize}

In this paper, the network architectures of the proposed and tested frameworks and the corresponding training strategy are presented in Section \ref{sec:method}. Section \ref{sec:brain} presents the experiments on brain tumor classification and Section \ref{sec:cardiac} presents the experiments on cardiac semantic level classification. Section \ref{sec:discussion} discusses the findings from the experiments and Section \ref{sec:conclusion} provides the conclusion.

\begin{figure*}[t]
    \centering
    \begin{minipage}[b]{1\linewidth}
      \centering
      \centerline{\includegraphics[width=1\linewidth]{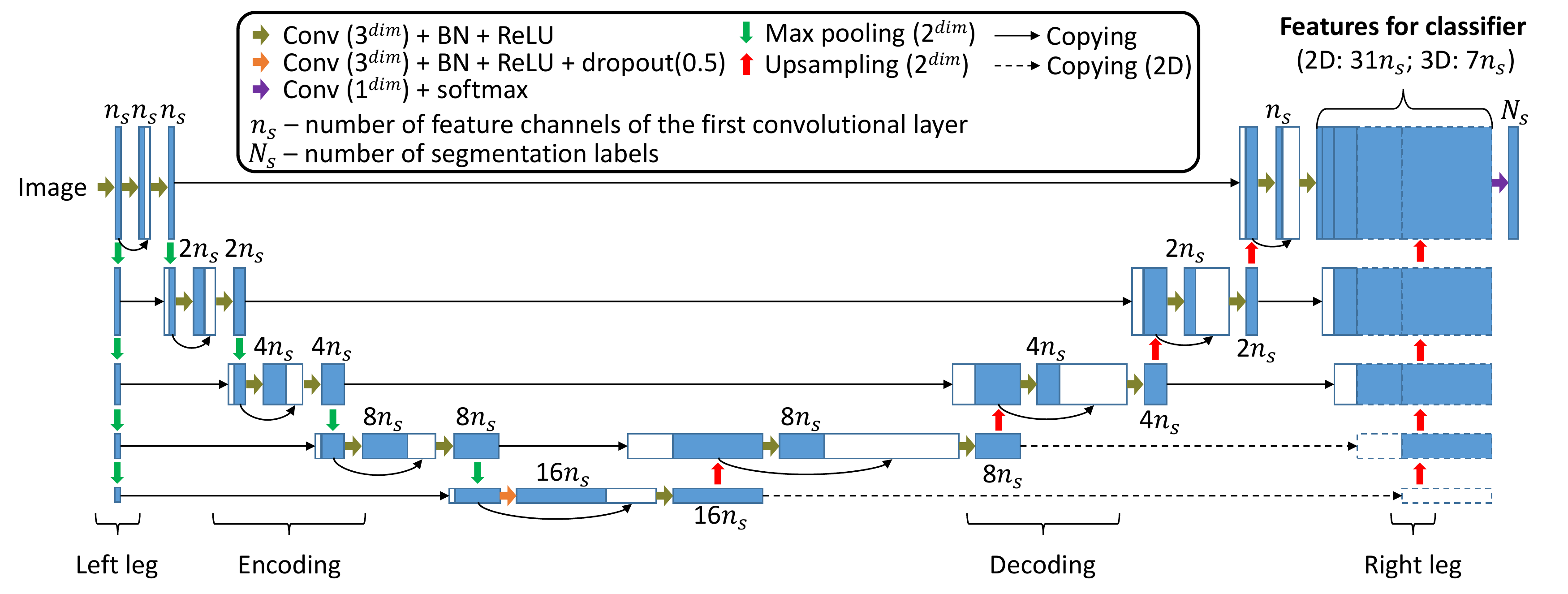}}
    \end{minipage}
    \caption{Segmentation network architecture. Blue blocks represent operation outputs and white blocks represent copied data. $dim \in \{2, 3\}$ is the input image dimension. At the right leg, dashed arrows and blocks are not used in 3D image analysis for computational feasibility. For the 2D cardiac CTA images, $n_s = 16$ and the number of features for the classifiers was $31n_s = 496$. For the 3D brain MR images, $n_s = 12$ and the number of features for the classifiers was $7n_s = 84$.}
    \label{fig:MNet}
\end{figure*}

\section{Methodology}
\label{sec:method}

Our classification framework includes a pre-trained segmentation network. The architecture of the segmentation network and the architectures of the proposed and tested classification frameworks are presented.

\subsection{Architecture of segmentation network}

The segmentation network in \cite{Conference:Mehta:ISBI2017} (M-Net) is modified to serve as a feature source for image classification (Fig. \ref{fig:MNet}). While the network was originally proposed for segmenting 3D images by stacking 2D segmentations, the modified architecture can be used for both 2D and 3D segmentation. The M-Net is modified from the U-Net \citep{Conference:Ronneberger:MICCAI2015}, which is a fully convolutional network that works with few training samples. Apart from the encoding and decoding paths of the U-Net, the M-Net also has two side paths (left and right legs) which provide functionality of deep supervision \citep{Conference:Lee:AISTATS2015}.

For our segmentation network, the M-Net is modified so that the number of feature channels of each convolutional layer systematically evolves with max pooling and upsampling. The number of channels is doubled after each pooling and halved after each upsampling. The encoding path consists of repetitions of two cascaded 3$^{dim}$ convolutional layers and a 2$^{dim}$ max pooling layer, with $dim \in \{2, 3\}$ the input image dimension. A dropout layer with 0.5 probability is introduced after the last pooling layer to reduce overfitting. The decoding path consists of repetitions of a 2$^{dim}$ upsampling layer, a concatenation with the corresponding feature channels from the encoding path, and two cascaded 3$^{dim}$ convolutional layers. Padding is used for each convolution to ensure segmentation image of the same size. Each 3$^{dim}$ convolution associates with batch normalization (BN) \citep{Conference:Ioffe:ICML2015} and rectified linear units (ReLU) \citep{Conference:Nair:ICML2010}, and skip connections are introduced between cascaded convolutional layers for better features learning \citep{Conference:Srivastava:NIPS2015}. Similar to \cite{Conference:Ronneberger:MICCAI2015}, we have four max pooling and four upsampling layers. For the two side paths, the left leg operates on the feature channels of the first convolutional layer with four 2$^{dim}$ max pooling layers, and the outputs are input to the corresponding encoding layers. The right leg consists of upsampling and concatenation that combine the outputs of the decoding layers. The feature channels of the last 3$^{dim}$ convolution and of the right leg are concatenated and input to the final 1$^{dim}$ convolutional layer with the softmax function to provide the segmentation probabilities. For computational feasibility, only parts of the outputs from the decoding path are used for the right leg for 3D segmentation.

This segmentation network can be pre-trained to provide useful features for classification. Compared with the last concatenation layer in the modified U-Net in \cite{Conference:Wong:MICCAI2017}, the right leg of this modified M-Net can provide more features (e.g. $3n_s$ in U-Net vs. $31n_s$ in M-Net for 2D images, Fig. \ref{fig:MNet}). Therefore, this modified M-Net is used as the feature source in our classification framework.

\begin{figure*}[t]
    \centering
    \begin{minipage}[b]{1\linewidth}
      \centering
      \centerline{\includegraphics[width=1\linewidth]{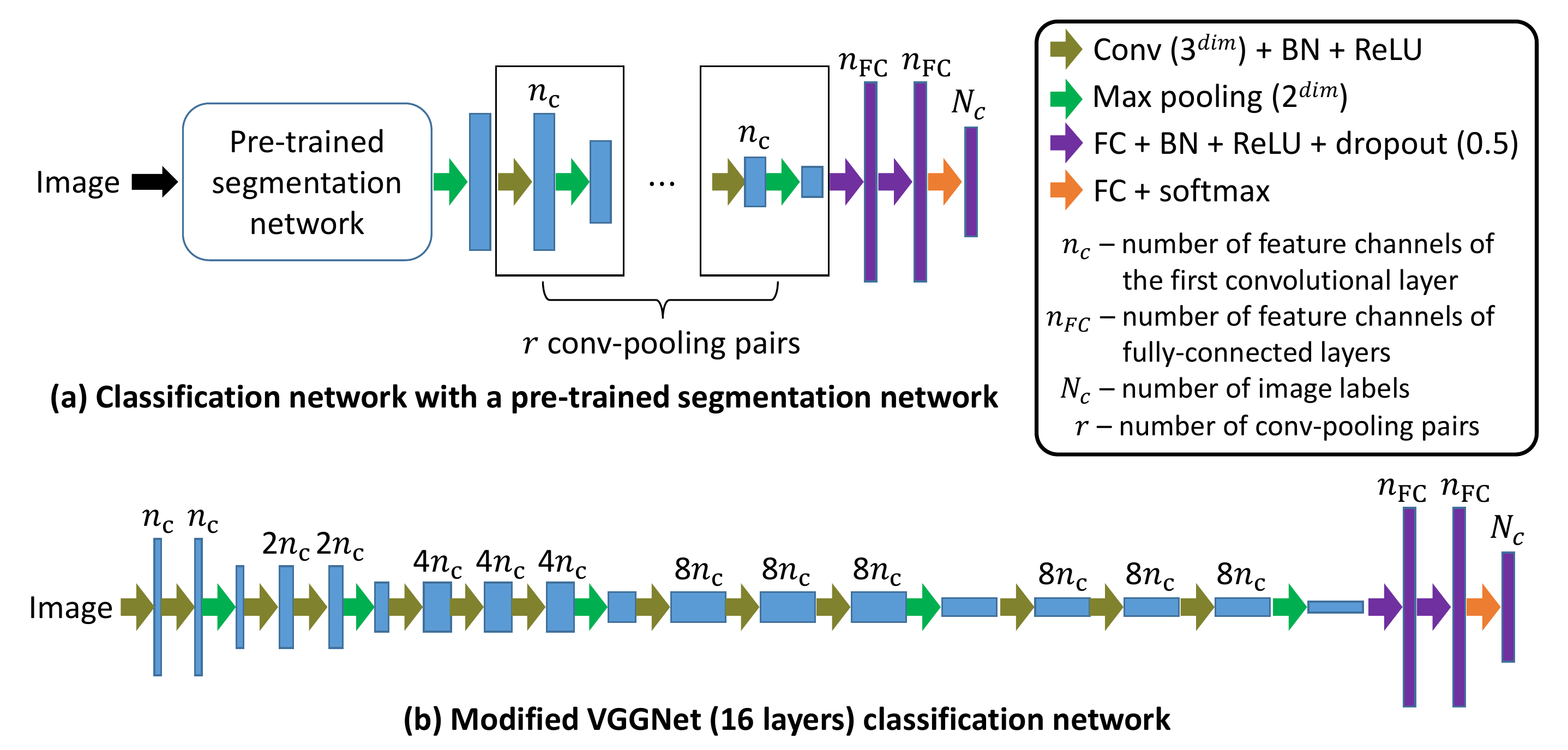}}
    \end{minipage}
    \caption{Classification network architectures. $dim \in \{2, 3\}$ is the input image dimension. (a) The proposed classification network architecture with a pre-trained segmentation network. For the 2D cardiac CTA images, $n_c = 16$, $n_{FC} = 100$, and $r = 3$. For the 3D brain MR images, $n_c = 32$, $n_{FC} = 200$, and $r = 4$. (b) A modified VGGNet architecture, which reproduces VGGNet when $n_c = 64$, $n_{FC}=4096$, and with batch normalization removed. For the 2D cardiac CTA images, $n_c = 16$ and $n_{FC} = 100$. For the 3D brain MR images, $n_c = 16$ and $n_{FC} = 200$.}
    \label{fig:ClassNets}
\end{figure*}

\subsection{Architectures of classification networks}

\subsubsection{Proposed architecture with a pre-trained segmentation network}
\label{sec:proposed}

The proposed classification network with a pre-trained segmentation network is shown Fig. \ref{fig:ClassNets}(a). Instead of combining with an existing complex architecture such as VGGNet \citep{Journal:Simonyan:arXiv2014}, we found that only a few convolutional layers with max pooling and fully-connected (FC) layers can provide high classification performance. Our classification network takes the features from a pre-trained segmentation network as inputs. The features are first downsampled by 2$^{dim}$ max pooling, followed by $r$ repetitions of a conv-pooling pair (a 3$^{dim}$ convolutional layer followed by a 2$^{dim}$ max pooling layer). Similar to the segmentation network, each 3$^{dim}$ convolution associates with BN and ReLU. The output from the final max pooling layer is flattened and passed to three cascaded FC layers. The first two FC layers associate with BN, ReLU, and dropout with 0.5 probability. The last FC layer with the softmax function provides the classification probabilities.

\subsubsection{Architecture of modified VGGNet}
\label{sec:VGG}

We compare the performance of our proposed classification network to a modified VGGNet (16 layers) architecture. We found in the early experiments that for tasks with relatively few classes compared to the 1000 classes in ImageNet, fewer feature channels can provide better performance with limited training data. Therefore, we modify the VGGNet architecture as shown in Fig. \ref{fig:ClassNets}(b), which is mostly the same as the original VGGNet architecture but using BN with the convolutional and FC layers. We also generalize the evolution of the numbers of channels of the convolutional layers with respect to the max pooling layers.

For 2D image classification, we also compare our framework to a classification network with features from an ImageNet pre-trained VGGNet model. To utilize the features, the FC layers of the pre-trained VGGNet model are replaced by those of the modified VGGNet architecture with $n_{FC} = 100$. Only the weights of the FC layers are modified during training. This allows us to compare our framework with a commonly used transfer learning strategy.

\subsection{Training strategy}
\label{sec:training_strategy}

Similar training strategies are used for both segmentation and classification networks. For the segmentation network, weighted cross entropy is used to compute the loss function as:
\begin{align}
\label{eq:loss}
    L = - \sum_\mathbf{x} w_{l(\mathbf{x})} \ln(p_{l(\mathbf{x})})
\end{align}
where $\mathbf{x}$ is the pixel position and $l(\mathbf{x})$ is the corresponding ground-truth label. $p_{l(\mathbf{x})}$ is the softmax probability of the channel corresponds to $l(\mathbf{x})$. $w_{l(\mathbf{x})}$ is the label weight for reducing the influences of more frequently seen labels. For the classification networks, the summation in (\ref{eq:loss}) is removed.

The label weights can be computed from the training data as:
\begin{align}
\label{eq:weights}
    w_l = \left(\frac{\sum_k f_k}{f_l}\right)^{0.5}
\end{align}
with $f_k$ the number of pixels with label $k$ in segmentation, or the number of images with label $k$ in classification. The exponent of 0.5 is used to avoid over penalty as the differences among label occurrences can be large. The stochastic optimization algorithm Adam is used for fast convergence \citep{Journal:Kingma:arXiv2014}. Glorot uniform initialization is used for both the convolutional and FC layers \citep{Conference:Glorot:AISTATS2010}.

With limited training data, image augmentation is important for learning invariant features and reducing overfitting. As computing realistic deformation of anatomical structures can be difficult and time consuming, while unrealistic deformation may introduce undesirable knowledge to the networks, we limit the image augmentation to rigid transformations, including rotation, shifting, zooming, and flipping (for symmetric structures such as the brain). In each epoch, each image has a probability (0.6 in our experiments) to be transformed by random transformations within given ranges, thus the number of augmented images is proportional to the number of epochs. Same setting is applied to all tested frameworks.

For the classification network with the ImageNet pre-trained VGGNet model, additional image pre-processing is required to comply with the input formats. Firstly, the images need to be resized to 224$\times$224$\times$3 to match the pre-training image size. Secondly, in medical images, the image intensity may not be within 0 and 255 as the ImageNet data do. In such situations, the intensity is normalized and shifted by -128 to resemble the pre-training condition. Contrast limited adaptive histogram equalization (CLAHE) is used for the normalization which also improves image contrast \citep{Conference:Zuiderveld:GG1994}. In fact, this normalization step may be unnecessary with the trainable FC layers, but we do this to ensure the network provides its best performance in the experiments.

\begin{table*}[t]
\caption{Semantic brain segmentation. Semantic labels and their corresponding Dice coefficients between prediction and ground truth on the testing data. CVL represents cerebellar vermal lobules. The average Dice coefficient was 69\%.}
\label{table:brain:semantic_dice}
\smallskip
\fontsize{6}{7}\selectfont
\centering
\begin{tabularx}{\linewidth}{XXXXX}
\toprule
Cerebral grey & 3rd ventricle & 4th ventricle & Brainstem & CVL I-V \\
\midrule
83\% & 59\% & 75\% & 80\% & 73\% \\
\midrule
CVL VI-VII & CVL VIII-X & Accumbens & Amygdala & Caudate \\
\midrule
61\% & 71\% & 45\% & 63\% & 63\% \\
\midrule
Cerebellar grey & Cerebellar white & Cerebral white & Hippocampus & Inf. lateral vent. \\
\midrule
82\% & 77\% & 86\% & 58\% & 46\% \\
\midrule
Lateral ventricle & Pallidum & Putamen & Thalamus \\
\midrule
84\% & 65\% & 71\% & 76\% \\
\bottomrule
\end{tabularx}
\end{table*}

\begin{figure*}[t]
    \scriptsize
    \centering
    \fbox{\noindent
    \fontsize{6}{7}\selectfont
    \begin{minipage}[b]{0.45\linewidth}
      \centering
      \begin{minipage}[b]{0.32\linewidth}
      \includegraphics[width=1\linewidth]{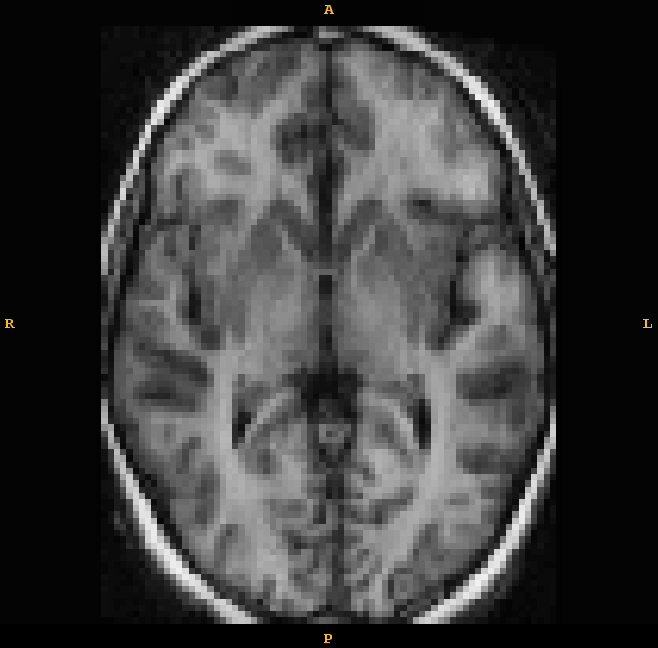} \\
      \centering{Image}
      \end{minipage}
      \begin{minipage}[b]{0.32\linewidth}
      \includegraphics[width=1\linewidth]{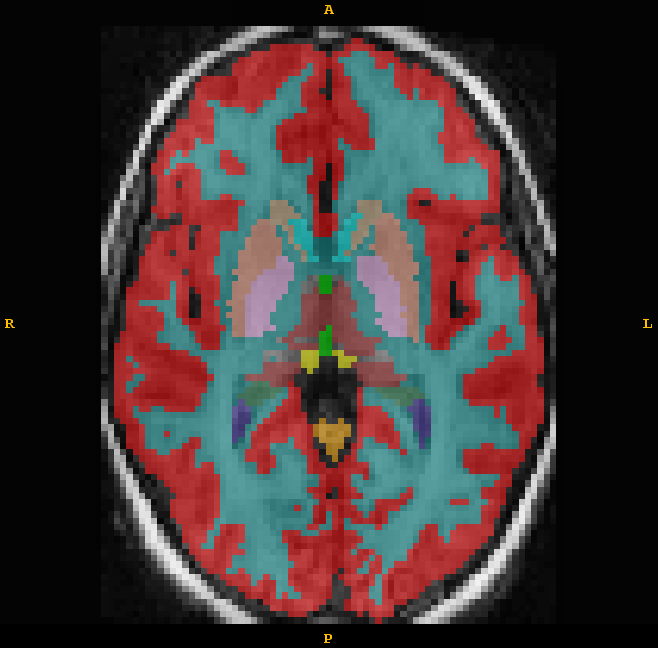}
      \centering{Ground}
      \end{minipage}
      \begin{minipage}[b]{0.32\linewidth}
      \includegraphics[width=1\linewidth]{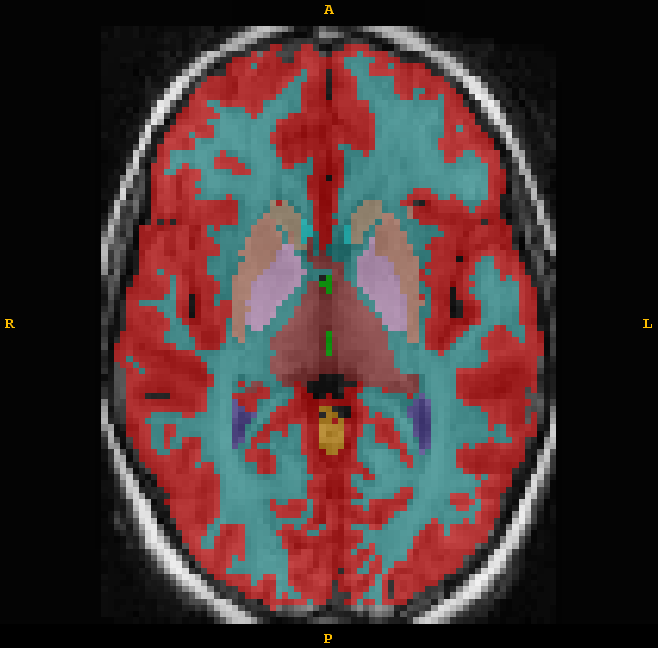}
      \centering{Seg}
      \end{minipage}
    \end{minipage}\noindent
    }
    \fbox{\noindent
    \fontsize{6}{7}\selectfont
    \begin{minipage}[b]{0.45\linewidth}
      \centering
      \begin{minipage}[b]{0.32\linewidth}
      \includegraphics[width=1\linewidth]{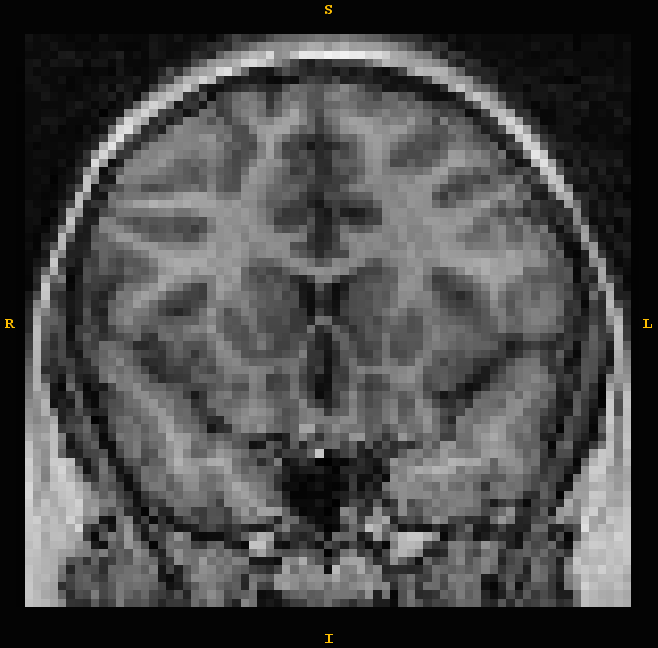} \\
      \centering{Image}
      \end{minipage}
      \begin{minipage}[b]{0.32\linewidth}
      \includegraphics[width=1\linewidth]{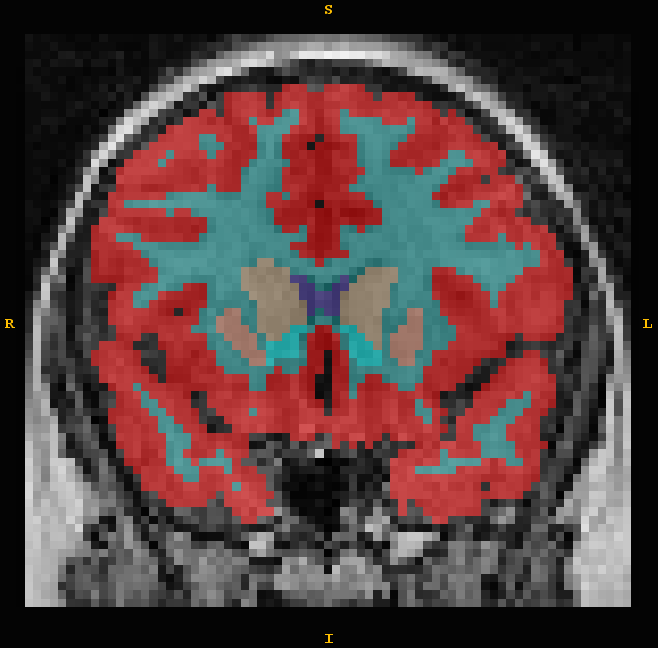}
      \centering{Ground}
      \end{minipage}
      \begin{minipage}[b]{0.32\linewidth}
      \includegraphics[width=1\linewidth]{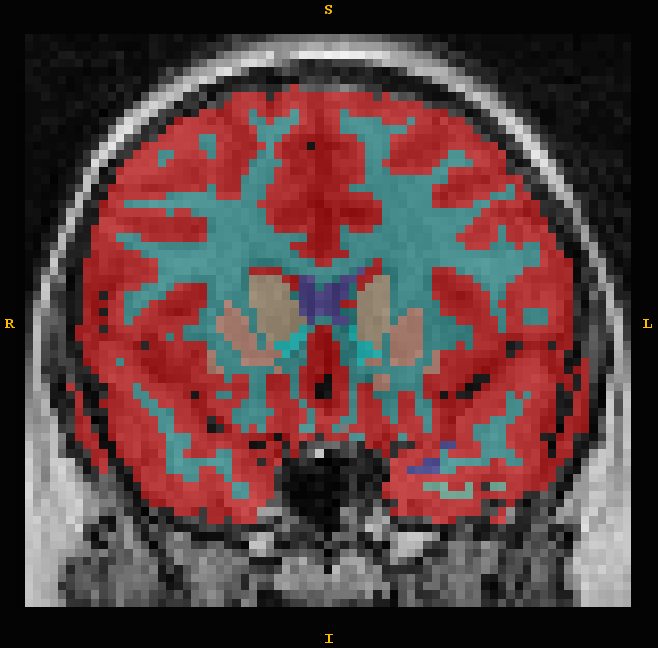}
      \centering{Seg}
      \end{minipage}
    \end{minipage}\noindent
    }
    \\
    \smallskip
    \centering{(a) Non-tumorous sample.}
    \\
    \centering
    \medskip
    \fbox{\noindent
    \fontsize{6}{7}\selectfont
    \begin{minipage}[b]{0.3\linewidth}
      \centering
      \begin{minipage}[b]{0.47\linewidth}
      \includegraphics[width=1\linewidth]{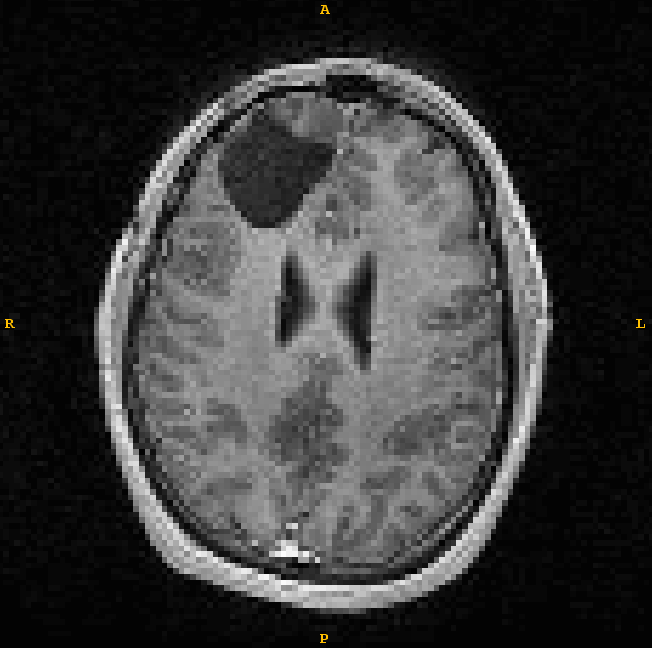}
      \centering{Image}
      \end{minipage}
      \begin{minipage}[b]{0.47\linewidth}
      \includegraphics[width=1\linewidth]{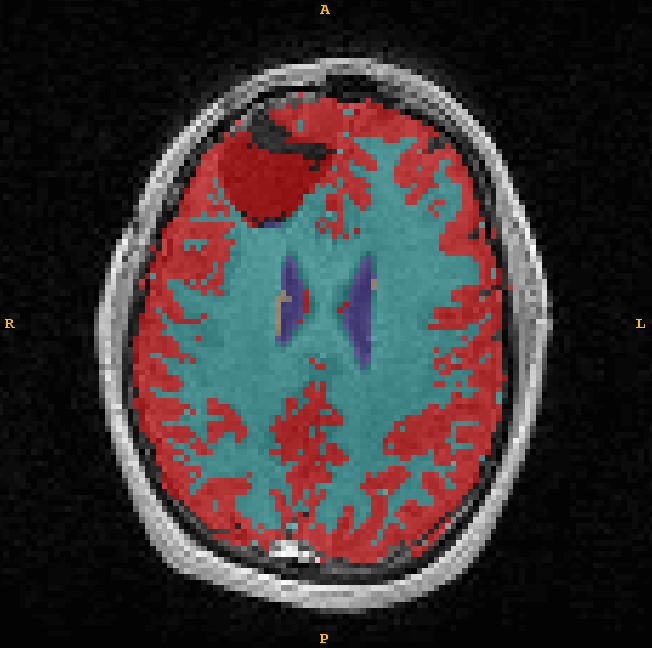}
      \centering{Seg}
      \end{minipage}
    \end{minipage}\noindent
    }
    \fbox{\noindent
    \fontsize{6}{7}\selectfont
    \begin{minipage}[b]{0.3\linewidth}
      \centering
      \begin{minipage}[b]{0.47\linewidth}
      \includegraphics[width=1\linewidth]{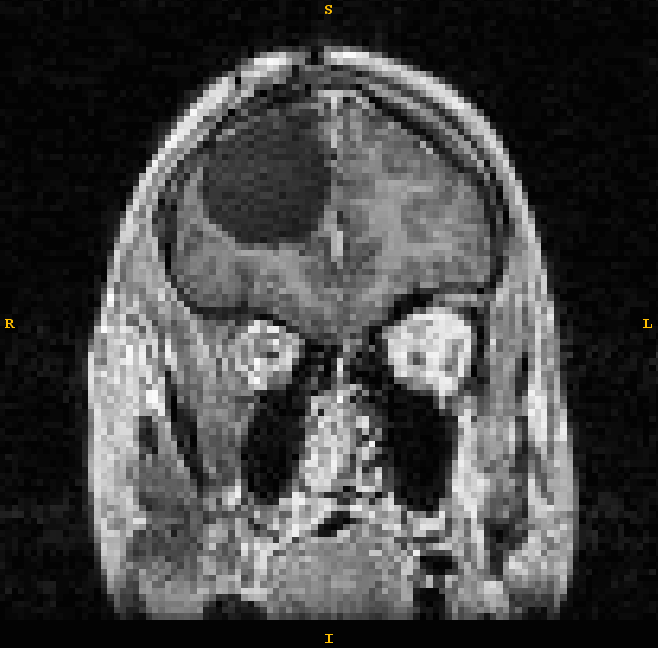}
      \centering{Image}
      \end{minipage}
      \begin{minipage}[b]{0.47\linewidth}
      \includegraphics[width=1\linewidth]{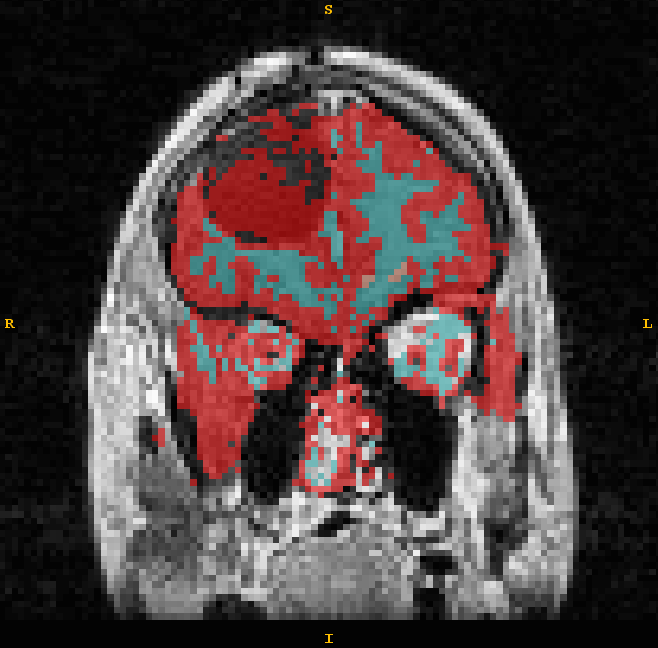}
      \centering{Seg}
      \end{minipage}
    \end{minipage}\noindent
    }
    \\
    \smallskip
    \centering{(b) Low-grade glioma.}
    \\
    \centering
    \medskip
    \fbox{\noindent
    \fontsize{6}{7}\selectfont
    \begin{minipage}[b]{0.3\linewidth}
      \centering
      \begin{minipage}[b]{0.47\linewidth}
      \includegraphics[width=1\linewidth]{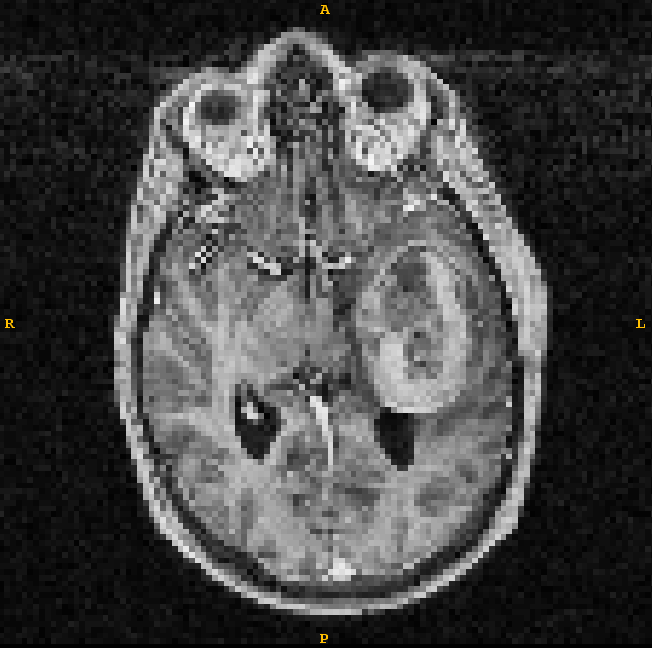}
      \centering{Image}
      \end{minipage}
      \begin{minipage}[b]{0.47\linewidth}
      \includegraphics[width=1\linewidth]{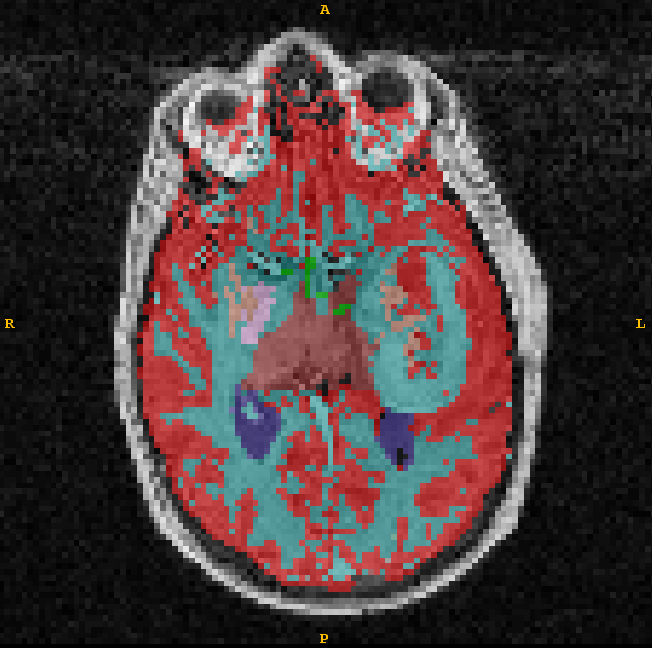}
      \centering{Seg}
      \end{minipage}
    \end{minipage}\noindent
    }
    \fbox{\noindent
    \fontsize{6}{7}\selectfont
    \begin{minipage}[b]{0.3\linewidth}
      \centering
      \begin{minipage}[b]{0.47\linewidth}
      \includegraphics[width=1\linewidth]{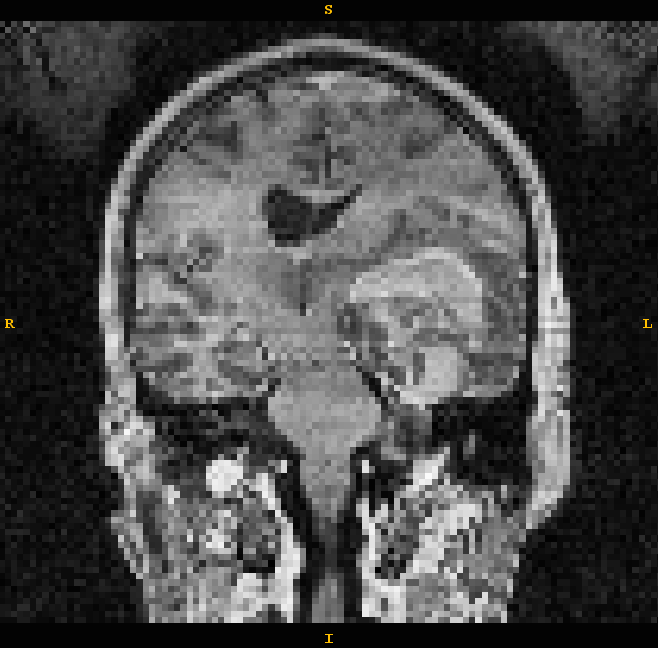}
      \centering{Image}
      \end{minipage}
      \begin{minipage}[b]{0.47\linewidth}
      \includegraphics[width=1\linewidth]{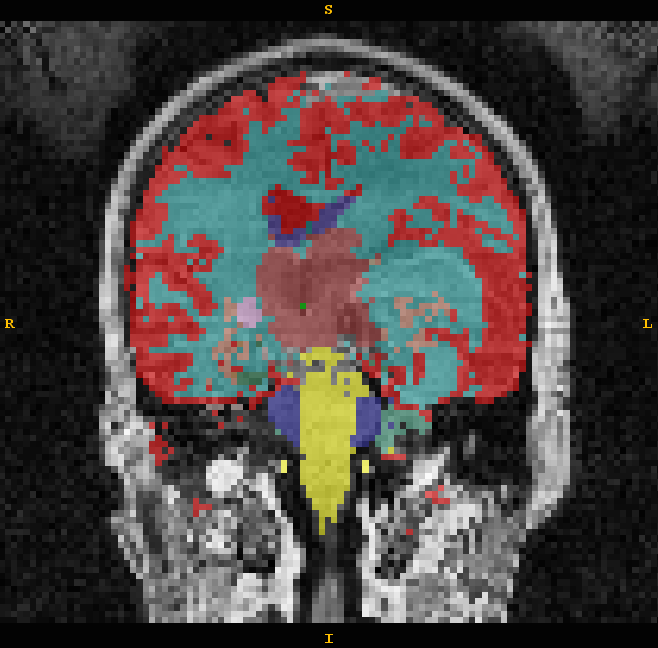}
      \centering{Seg}
      \end{minipage}
    \end{minipage}\noindent
    }
    \\
    \smallskip
    \centering{(c) Glioblastoma.}
    \caption{Semantic brain segmentation. (a) An unseen non-tumorous sample from the segmentation testing data in axial and coronal views. (b) and (c) Samples with low-grade glioma and glioblastoma (no ground truth).}
    \label{fig:brain:seg}
\end{figure*}

\begin{figure*}[t]
    \scriptsize
    \centering
    \medskip
    \fbox{\noindent
    \fontsize{6}{7}\selectfont
    \begin{minipage}[b]{0.3\linewidth}
      \centering
      \begin{minipage}[b]{0.47\linewidth}
      \includegraphics[width=1\linewidth]{brain_LGG_24_A_img}
      \centering{Image}
      \end{minipage}
      \begin{minipage}[b]{0.47\linewidth}
      \includegraphics[width=1\linewidth]{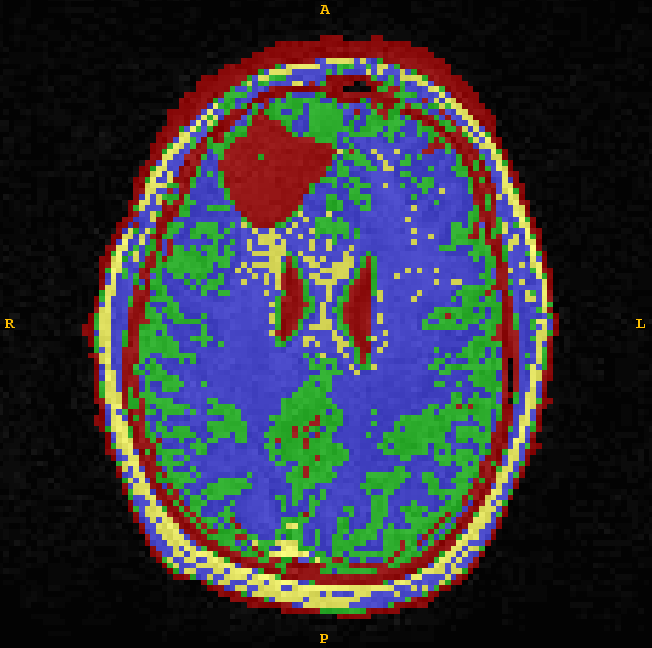}
      \centering{Seg}
      \end{minipage}
    \end{minipage}\noindent
    }
    \fbox{\noindent
    \fontsize{6}{7}\selectfont
    \begin{minipage}[b]{0.3\linewidth}
      \centering
      \begin{minipage}[b]{0.47\linewidth}
      \includegraphics[width=1\linewidth]{brain_LGG_24_C_img}
      \centering{Image}
      \end{minipage}
      \begin{minipage}[b]{0.47\linewidth}
      \includegraphics[width=1\linewidth]{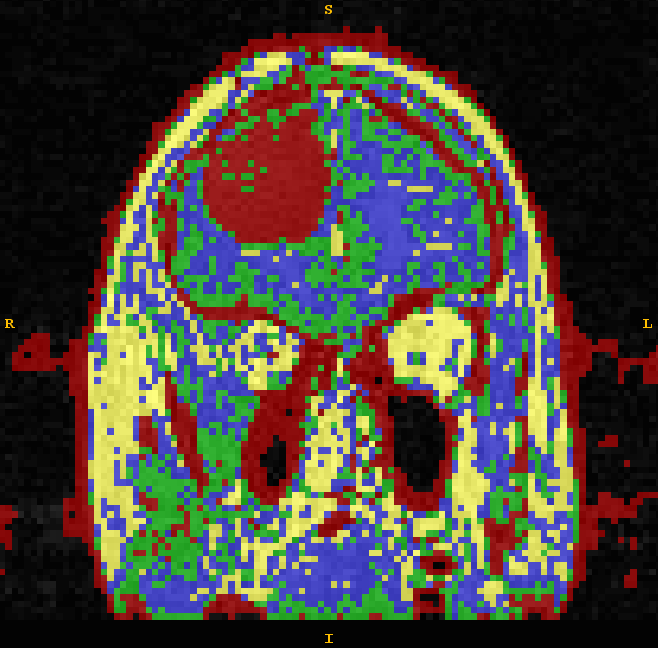}
      \centering{Seg}
      \end{minipage}
    \end{minipage}\noindent
    }
    \\
    \smallskip
    \centering{(a) Low-grade glioma.}
    \\
    \centering
    \medskip
    \fbox{\noindent
    \fontsize{6}{7}\selectfont
    \begin{minipage}[b]{0.3\linewidth}
      \centering
      \begin{minipage}[b]{0.47\linewidth}
      \includegraphics[width=1\linewidth]{brain_GBM_40_A_img}
      \centering{Image}
      \end{minipage}
      \begin{minipage}[b]{0.47\linewidth}
      \includegraphics[width=1\linewidth]{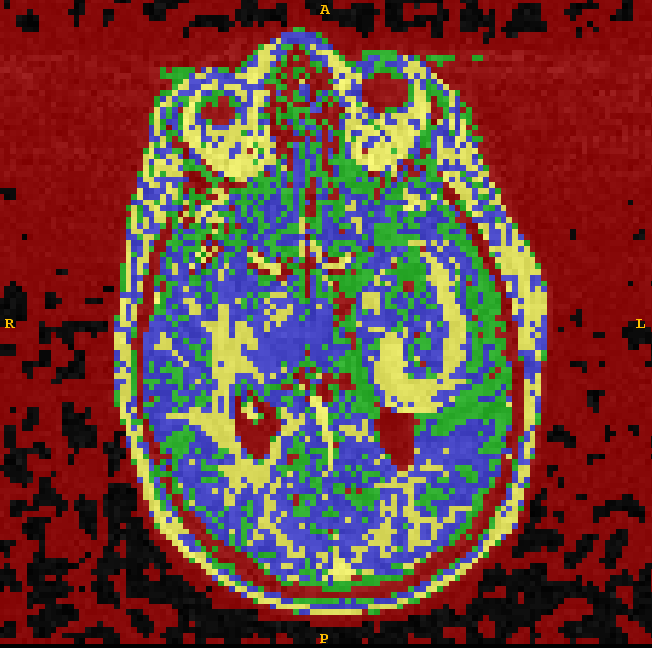}
      \centering{Seg}
      \end{minipage}
    \end{minipage}\noindent
    }
    \fbox{\noindent
    \fontsize{6}{7}\selectfont
    \begin{minipage}[b]{0.3\linewidth}
      \centering
      \begin{minipage}[b]{0.47\linewidth}
      \includegraphics[width=1\linewidth]{brain_GBM_40_C_img}
      \centering{Image}
      \end{minipage}
      \begin{minipage}[b]{0.47\linewidth}
      \includegraphics[width=1\linewidth]{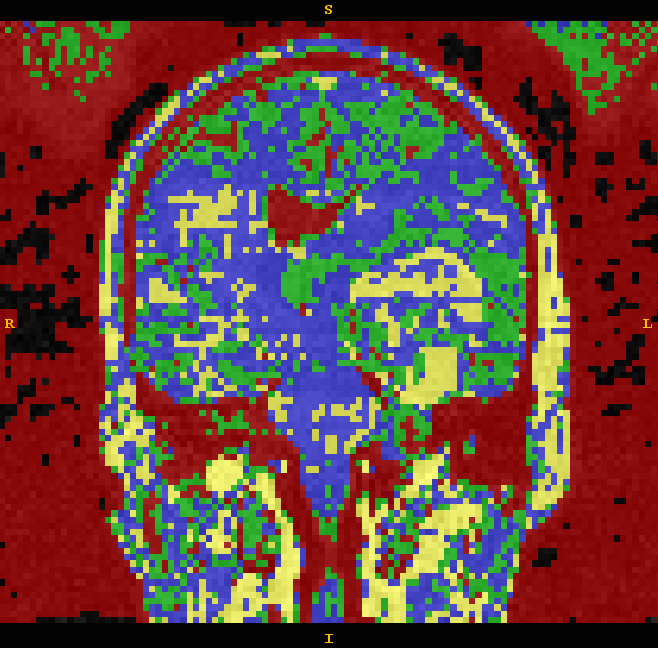}
      \centering{Seg}
      \end{minipage}
    \end{minipage}\noindent
    }
    \\
    \smallskip
    \centering{(b) Glioblastoma.}
    \caption{Non-semantic brain segmentation with low-grade glioma and glioblastoma in axial and coronal views (no ground truth).}
    \label{fig:brain:seg_nonsemantic}
\end{figure*}

\begin{figure*}[t]
    \scriptsize
    \centering
    \begin{minipage}[b]{0.32\linewidth}
      \centering
      \begin{minipage}[b]{\linewidth}
      \includegraphics[width=1\linewidth]{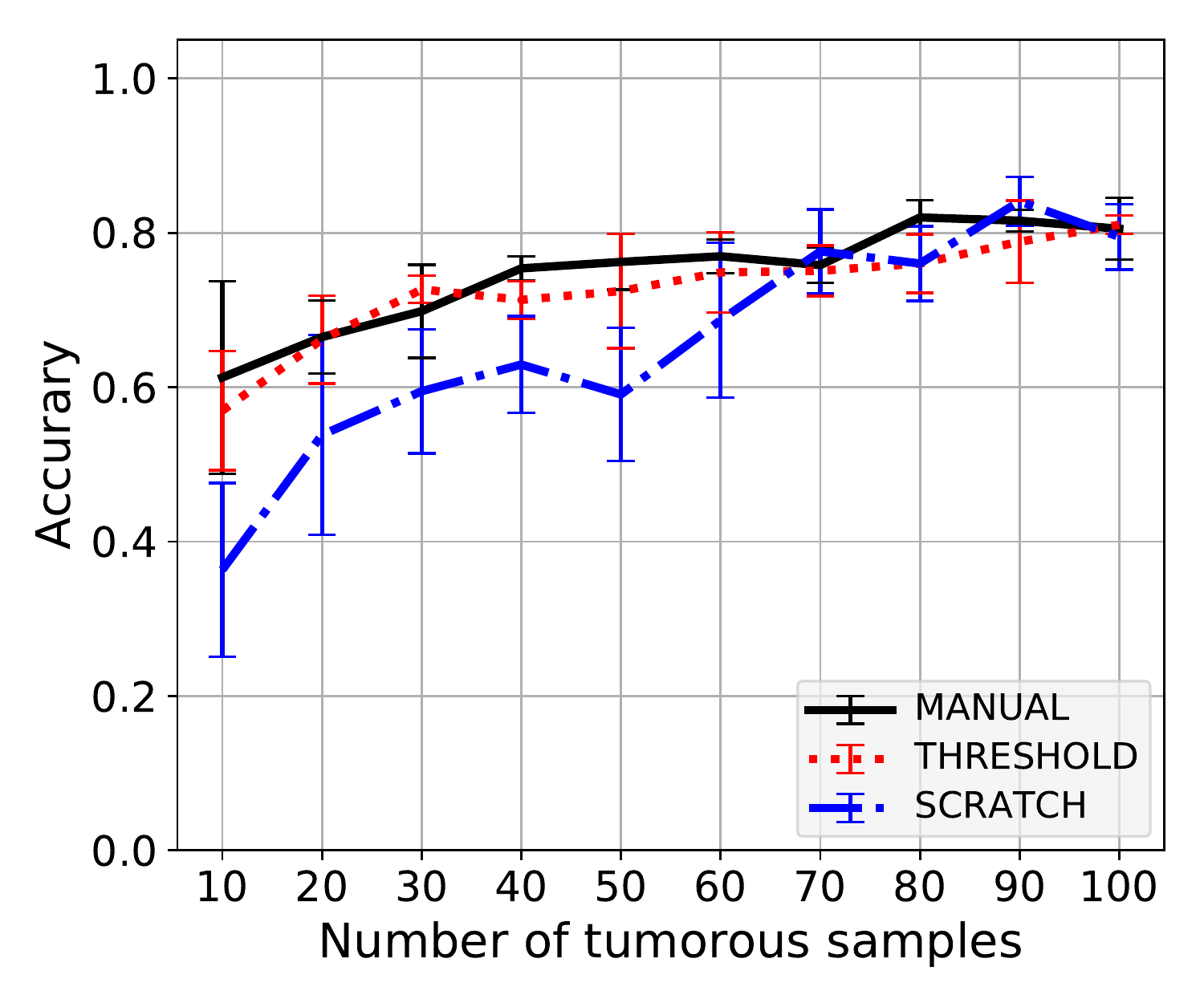}
      \centering{(a) Accuracy.}
      \end{minipage}
    \end{minipage}
    \begin{minipage}[b]{0.32\linewidth}
      \centering
      \begin{minipage}[b]{\linewidth}
      \includegraphics[width=1\linewidth]{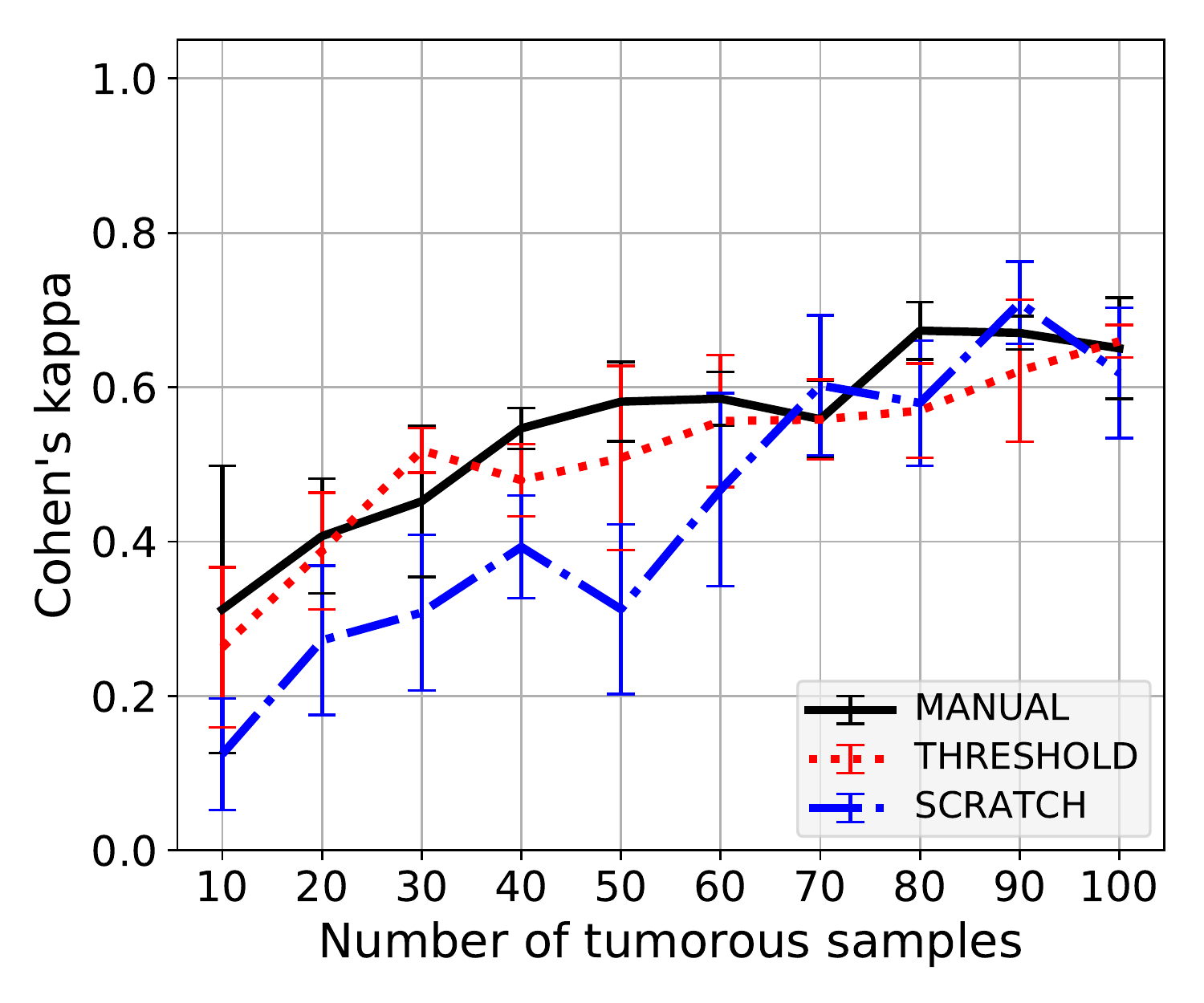}
      \centering{(b) Cohen's kappa.}
      \end{minipage}
    \end{minipage}
    \\
    \centering
    \begin{minipage}[b]{\linewidth}
      \includegraphics[width=1\linewidth]{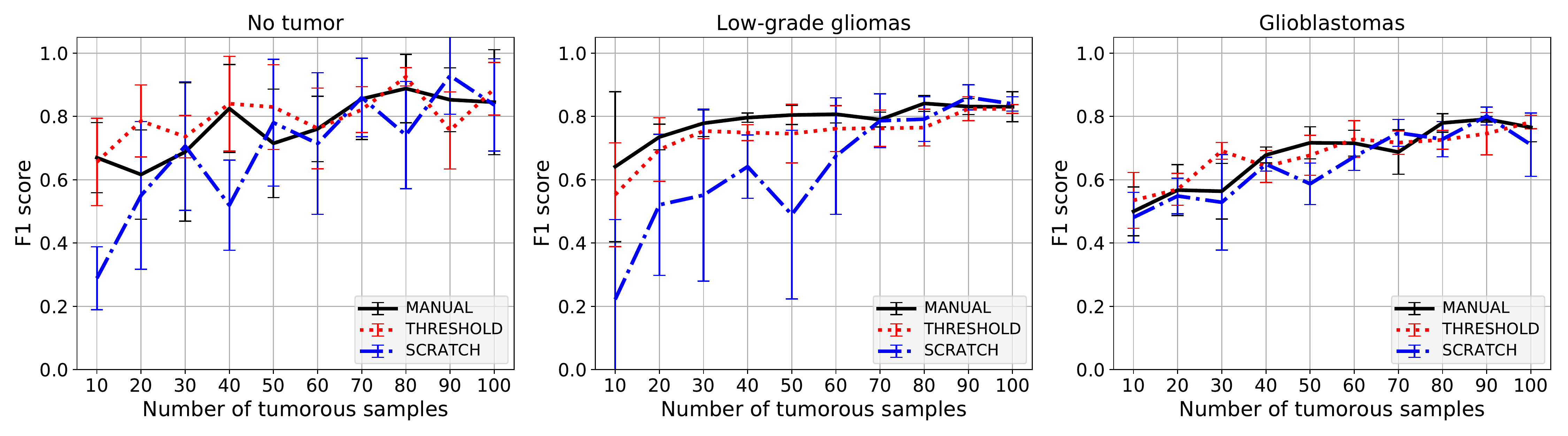}
      \centering{(c) F1 score.}
    \end{minipage}
    \caption{Brain tumor classification. Results of the tested frameworks (Section \ref{sec:frameworks}). Each data point corresponds to five repeated experiments, with the lines representing the average values and the error bars representing the standard deviations. The $x$-axis represents the number of tumorous training samples (five to 50 for each class), and the number of non-tumorous samples was fixed at 11. The accuracies and Cohen's kappas were computed from all testing samples, while the F1 scores were computed for each class.}
    \label{fig:brain:classification}
\end{figure*}

\subsection{Implementation}

The Python deep learning library Keras was used with the TensorFlow backend \citep{Misc:Chollet:Keras2017}. The DIGITS DevBox of NVIDIA was used for the experiments, which has four TITAN X GPUs with 12 GB of memory per GPU. Only one GPU was used for each experiment.

\subsection{Tested frameworks for image classification}
\label{sec:frameworks}

With the black-box nature of CNNs, it is difficult to perform fundamental or theoretical analysis. Although fundamental analysis such as that in \cite{Conference:Bengio:ICML2009} exists, it is very challenging to develop a similar analysis for our framework. Therefore, we demonstrate the advantages of our proposed framework through comparisons with other frameworks on different problems, organs, and image modalities. The tested frameworks include:

\begin{itemize}
  \item \textbf{MANUAL}: the classification network in Fig. \ref{fig:ClassNets}(a) (Section \ref{sec:proposed}) with the segmentation network pre-trained on samples with manual segmentation by radiologists.
  \item \textbf{THRESHOLD}: the classification network in Fig. \ref{fig:ClassNets}(a) (Section \ref{sec:proposed}) with the segmentation network pre-trained on samples with segmentation obtained by intensity thresholding, i.e. $k$-means clustering.
  \item \textbf{VGG}: the classification network in Fig. \ref{fig:ClassNets}(b) with the convolutional features from the ImageNet pre-trained VGGNet model (Section \ref{sec:VGG}). As the 3D counterpart of the ImageNet pre-trained VGGNet model does not exist, this framework was only tested on 2D image classification.
  \item \textbf{SCRATCH}: the classification network in Fig. \ref{fig:ClassNets}(b) (Section \ref{sec:VGG}) trained from scratch without any pre-trained features.
\end{itemize}
The MANUAL framework is the optimal configuration of our proposed framework with pre-trained features from semantic segmentation. As manual segmentation can be expensive and time consuming, we use the THRESHOLD framework to study the use of non-semantic labels from image thresholding. The VGG framework represents a commonly used transfer learning technique \citep{Conference:Yosinski:NIPS2014,Journal:Shin:TMI2016}. The SCRATCH framework acts as a baseline for comparison. These frameworks were tested on tumor classification and cardiac semantic level classification.

\section{Tumor classification on brain MR images}
\label{sec:brain}

Our goal is to distinguish among no tumor, low-grade gliomas (LGG), and glioblastomas. Gliomas are brain tumors which can be classified into Grade I (least malignant) through Grade IV (most malignant) using the World Health Organization classification system \citep{Journal:Louis:AN2007}. LGG consist of Grade I and II gliomas while glioblastomas are Grade IV \citep{Journal:van:Lancet2011,Journal:Stark:CNN2012}. In T1-weighted MR images, LGG are often homogeneous with low image intensity, while glioblastomas are more heterogeneous and contrast enhanced (Fig. \ref{fig:brain:seg} (b) and (c)).

\subsection{Segmentation network pre-training}

\subsubsection{Data and training}
\label{sec:brain:data_training}

A dataset of 43 3D images from patients without brain tumors was neuroanatomically labeled to provide the training and testing data for the segmentation network. The images were produced by the T1-weighted, magnetization-prepared rapid gradient-echo (MP-RAGE) pulse sequence which provides good tissue contrast for anatomical segmentation. They were manually segmented by highly trained experts with the results reviewed by a consulting neuroanatomist. The 19 anatomical labels used are shown in Table \ref{table:brain:semantic_dice}.

From the dataset, 21 volumes were used for training and 22 were used for testing. All volumes were resized to 128$\times$128$\times$128 pixels. Different from the Hounsfield unit in CT, the intensity values of MR images do not directly characterize material properties, and the variations can be large with different settings. Therefore, we normalized the image intensity between 0 and 255 using CLAHE for consistency. As mentioned in Section \ref{sec:training_strategy}, image augmentation was used to learn invariant features and reduce overfitting. The segmentation network was trained with the batch size of 1, learning rate of 10$^{-3}$, and 100 epochs. We used $n_s = 12$ (Fig. \ref{fig:MNet}) as this produced a relatively small model that can fit in a GPU with 12 GB of memory during training. $N_s = 20$ with the background included. These hyperparameters were chosen as they provided the best performance among the tested parameters.

\subsubsection{Semantic segmentation results}

As the segmentation performance is not the main concern of this paper, only the results of the trained segmentation network that was used by the classification network are presented. The Dice coefficient between the prediction and the ground truth of each semantic label was computed for each volume, and the average values are reported in Table \ref{table:brain:semantic_dice}.

Some structures with Dice coefficients $\leq$ 63\%, such as the nucleus accumbens, amygdala, and inferior lateral ventricle, were small structures which were more prone to segmentation error. Structures which were relatively large or with simpler patterns, such as the cerebral grey and white matters, brainstem, and lateral ventricle, were well segmented with Dice coefficients $\geq$ 80\%. The average Dice coefficient was 69\%.

Fig. \ref{fig:brain:seg} shows examples of segmentation on unseen images. For the segmentation on the testing data (Fig. \ref{fig:brain:seg}(a)), the results were generally consistent with the ground truths but with some missing details. On the other hand, as our goal is to apply the features from the trained segmentation network for tumor classification, we also show some segmentation results on images with LGG and glioblastoma in Fig. \ref{fig:brain:seg}(b) and (c). As the semantic labels did not account for tumors, they were not segmented. Nevertheless, the segmented structures around the tumors were visually distinguishable from their non-tumorous counterparts, and the overall segmentations were less symmetric. Therefore, useful information to distinguish between non-tumorous and tumorous cases was embedded in the feature channels. Note that we have 7$n_s$ = 84 features for the classification (Fig. \ref{fig:MNet}).

\subsubsection{Non-semantic segmentation results}

As mentioned in Section \ref{sec:frameworks}, we also tested a framework with a segmentation network trained from non-semantic labels (THRESHOLD). The same images and training strategy in Section \ref{sec:brain:data_training} were used to train this segmentation network, with the five ground-truth labels provided by the $k$-means method. The number of classes was chosen so that segmented key structures such as grey and white matters were visually distinguishable. The segmentation results on unseen tumorous images are shown in Fig. \ref{fig:brain:seg_nonsemantic}. Although the segmentations were noisy, similar to the semantic segmentation, the segmented structures around the tumors were partially distinguishable.

\subsection{Tumor classification}

\subsubsection{Data and training}

The tumorous images were obtained from The Cancer Genome Atlas (TCGA) from patients with LGG and glioblastomas \citep{Misc:TCGA2017}. As the MR images for training the segmentation network were produced by the MP-RAGE pulse sequence, images from the TCGA dataset produced by the MP-RAGE or spoiled gradient recalled (SPGR) pulse sequences which produce similar tissue contrast were used. This provided 155 and 125 3D images for LGG and glioblastomas, respectively. For both LGG and glioblastomas, we used 50 samples for training and the rest for testing. This resulted in 50 and 105 images for training and testing for LGG, and 50 and 75 for training and testing for glioblastomas.

The testing data of the segmentation network provided the non-tumorous samples. By splitting the dataset in half, 11 images were produced for each training and testing.

Similar to the segmentation network, all images were resized to 128$\times$128$\times$128 pixels, and the image intensity was normalized between 0 and 255 using CLAHE. Image augmentation was used to learn invariant features and reduce overfitting. The classification networks were trained with the batch size of 4, learning rate of 10$^{-4}$, and 50 epochs. For the MANUAL and THRESHOLD frameworks (Fig. \ref{fig:ClassNets}(a)), $n_c = 32$, $n_{FC} = 200$, and $r = 4$. For the SCRATCH framework, $n_c = 16$ and $n_{FC} = 200$. $N_c = 3$ for the three classes of no tumor, LGG, and glioblastomas. These hyperparameters were chosen so that all frameworks converged and achieved the best performances when all training data were used. To study the performance with limited samples, the tumorous training samples of each of the LGG and glioblastomas were gradually increased from five to 50 with strides of five (i.e. 10 to 100 with strides of 10 for all tumorous samples). All non-tumorous training samples (11 3D images) were used given the small number. Five repetitions were performed on randomized splits of training and testing data. Identical data and setting were applied to all frameworks.

\subsubsection{Classification results}

Fig. \ref{fig:brain:classification} shows the results of the tested frameworks. Their performances were assessed by the classification accuracy, Cohen's kappa, and F1 score. The Cohen's kappa ($\in$[-1, 1]) measures the inter-rater agreement between the ground-truth and predicted classes. The F1 score ($\in$[0, 1]) is the harmonic average of the recall and precision. For each framework, the accuracy and Cohen's kappa were computed for the overall performance, while the F1 scores were computed for each class.

The accuracies and Cohen's kappas show that the frameworks with a pre-trained segmentation network (MANUAL and THRESHOLD) outperformed the SCRATCH framework when the number of tumorous samples was small. All frameworks started to have similar performances at 70 tumorous samples. The standard deviations of the SCRATCH framework were larger than those of the other frameworks in general. The MANUAL framework performed slightly better than the THRESHOLD framework, with the average absolute differences in accuracy and Cohen's kappa as 3\% and 5\%, respectively. On the contrary, the average absolute differences between the MANUAL and the SCRATCH frameworks in accuracy and Cohen's kappa were 10\% and 12\%, respectively. The MANUAL framework converged at 80 tumorous samples with the accuracy and Cohen's kappa as 82\% and 67\%, respectively.

The F1 scores of each class provide further insights. The differences were larger in the no tumor and LGG classes, for which the MANUAL and THRESHOLD frameworks performed better than the SCRATCH framework in general. For the no tumor class, although the number of normal training samples did not change, the F1 scores improved with the number of tumorous samples, though the fluctuations and standard deviations were relatively large. The MANUAL and THRESHOLD frameworks performed well with LGG even with small numbers of training samples, for example, the MANUAL framework had F1 score of 80\% with 40 tumorous samples. In contrast, the SCRATCH framework had relatively low scores before 70 tumorous samples with much larger fluctuations and standard deviations. For glioblastomas, the performances of all frameworks were similar.

\begin{table*}[t]
\caption{Semantic heart segmentation. Semantic labels and their corresponding Dice coefficients between prediction and ground truth on the testing data. PA stands for pulmonary artery. The average Dice coefficient was 78\%.}
\label{table:cardiac:semantic_dice}
\smallskip
\fontsize{6}{7}\selectfont
\centering
\begin{tabularx}{\linewidth}{XXXX}
\toprule
Sternum & Ascending aorta & Descending aorta & Right PA \\
\midrule
69\% & 78\% & 81\% & 81\% \\
\midrule
Vertebrae & Right atrium & Left atrium & Right ventricle \\
\midrule
74\% & 85\% & 88\% & 81\% \\
\midrule
Left ventricle & Myocardium & Left PA & PA trunk \\
\midrule
80\% & 87\% & 81\% & 82\% \\
\midrule
Aortic root & Aortic arch & Superior vena cava & Inferior vena cava \\
\midrule
62\% & 76\% & 79\% & 62\% \\
\bottomrule
\end{tabularx}
\end{table*}

\begin{figure*}[t]
    \scriptsize
    \centering
    \fbox{\noindent
    \fontsize{6}{7}\selectfont
    \begin{minipage}[b][0.14\linewidth]{0.31\linewidth}
      \centering{Thoracic inlet}
      \\
      \smallskip
      \begin{minipage}[b]{0.32\linewidth}
      \includegraphics[width=1\linewidth]{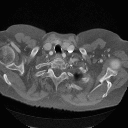} \\
      \centering{Image}
      \end{minipage}
      \begin{minipage}[b]{0.32\linewidth}
      \includegraphics[width=1\linewidth]{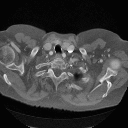}
      \centering{Ground}
      \end{minipage}
      \begin{minipage}[b]{0.32\linewidth}
      \includegraphics[width=1\linewidth]{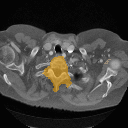}
      \centering{Seg}
      \end{minipage}
    \end{minipage}\noindent
    }
    \fbox{\noindent
    \fontsize{6}{7}\selectfont
    \begin{minipage}[b][0.14\linewidth]{0.31\linewidth}
      \centering{Lung apex}
      \\
      \smallskip
      \begin{minipage}[b]{0.32\linewidth}
      \includegraphics[width=1\linewidth]{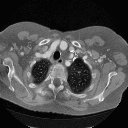} \\
      \centering{Image}
      \end{minipage}
      \begin{minipage}[b]{0.32\linewidth}
      \includegraphics[width=1\linewidth]{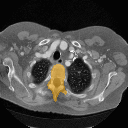}
      \centering{Ground}
      \end{minipage}
      \begin{minipage}[b]{0.32\linewidth}
      \includegraphics[width=1\linewidth]{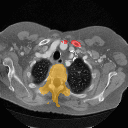}
      \centering{Seg}
      \end{minipage}
    \end{minipage}\noindent
    }
    \fbox{\noindent
    \fontsize{6}{7}\selectfont
    \begin{minipage}[b][0.14\linewidth]{0.31\linewidth}
      \centering{Great vessels origin}
      \\
      \smallskip
      \begin{minipage}[b]{0.32\linewidth}
      \includegraphics[width=1\linewidth]{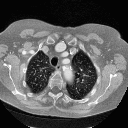} \\
      \centering{Image}
      \end{minipage}
      \begin{minipage}[b]{0.32\linewidth}
      \includegraphics[width=1\linewidth]{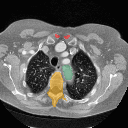}
      \centering{Ground}
      \end{minipage}
      \begin{minipage}[b]{0.32\linewidth}
      \includegraphics[width=1\linewidth]{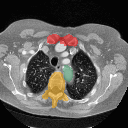}
      \centering{Seg}
      \end{minipage}
    \end{minipage}\noindent
    }
    \\
    \centering
    \fbox{\noindent
    \fontsize{6}{7}\selectfont
    \begin{minipage}[b][0.14\linewidth]{0.31\linewidth}
      \centering{Aortic arch}
      \\
      \smallskip
      \begin{minipage}[b]{0.32\linewidth}
      \includegraphics[width=1\linewidth]{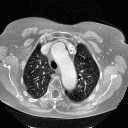} \\
      \centering{Image}
      \end{minipage}
      \begin{minipage}[b]{0.32\linewidth}
      \includegraphics[width=1\linewidth]{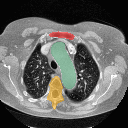}
      \centering{Ground}
      \end{minipage}
      \begin{minipage}[b]{0.32\linewidth}
      \includegraphics[width=1\linewidth]{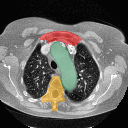}
      \centering{Seg}
      \end{minipage}
    \end{minipage}\noindent
    }
    \fbox{\noindent
    \fontsize{6}{7}\selectfont
    \begin{minipage}[b][0.14\linewidth]{0.31\linewidth}
      \centering{Ascending/descending aorta}
      \\
      \smallskip
      \begin{minipage}[b]{0.32\linewidth}
      \includegraphics[width=1\linewidth]{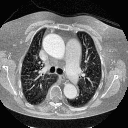} \\
      \centering{Image}
      \end{minipage}
      \begin{minipage}[b]{0.32\linewidth}
      \includegraphics[width=1\linewidth]{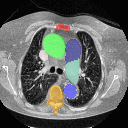}
      \centering{Ground}
      \end{minipage}
      \begin{minipage}[b]{0.32\linewidth}
      \includegraphics[width=1\linewidth]{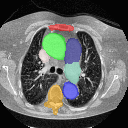}
      \centering{Seg}
      \end{minipage}
    \end{minipage}\noindent
    }
    \fbox{\noindent
    \fontsize{6}{7}\selectfont
    \begin{minipage}[b][0.14\linewidth]{0.31\linewidth}
      \centering{Pulmonary trunk}
      \\
      \smallskip
      \begin{minipage}[b]{0.32\linewidth}
      \includegraphics[width=1\linewidth]{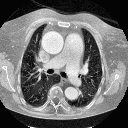} \\
      \centering{Image}
      \end{minipage}
      \begin{minipage}[b]{0.32\linewidth}
      \includegraphics[width=1\linewidth]{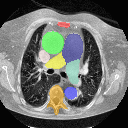}
      \centering{Ground}
      \end{minipage}
      \begin{minipage}[b]{0.32\linewidth}
      \includegraphics[width=1\linewidth]{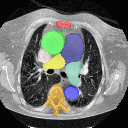}
      \centering{Seg}
      \end{minipage}
    \end{minipage}\noindent
    }
    \\
    \centering
    \fbox{\noindent
    \fontsize{6}{7}\selectfont
    \begin{minipage}[b][0.14\linewidth]{0.31\linewidth}
      \centering{Aortic valve/root}
      \\
      \smallskip
      \begin{minipage}[b]{0.32\linewidth}
      \includegraphics[width=1\linewidth]{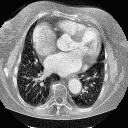} \\
      \centering{Image}
      \end{minipage}
      \begin{minipage}[b]{0.32\linewidth}
      \includegraphics[width=1\linewidth]{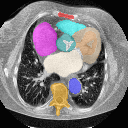}
      \centering{Ground}
      \end{minipage}
      \begin{minipage}[b]{0.32\linewidth}
      \includegraphics[width=1\linewidth]{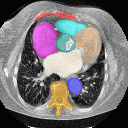}
      \centering{Seg}
      \end{minipage}
    \end{minipage}\noindent
    }
    \fbox{\noindent
    \fontsize{6}{7}\selectfont
    \begin{minipage}[b][0.14\linewidth]{0.31\linewidth}
      \centering{Axial 4-chamber view}
      \\
      \smallskip
      \begin{minipage}[b]{0.32\linewidth}
      \includegraphics[width=1\linewidth]{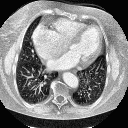} \\
      \centering{Image}
      \end{minipage}
      \begin{minipage}[b]{0.32\linewidth}
      \includegraphics[width=1\linewidth]{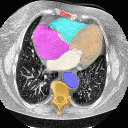}
      \centering{Ground}
      \end{minipage}
      \begin{minipage}[b]{0.32\linewidth}
      \includegraphics[width=1\linewidth]{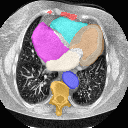}
      \centering{Seg}
      \end{minipage}
    \end{minipage}\noindent
    }
    \fbox{\noindent
    \fontsize{6}{7}\selectfont
    \begin{minipage}[b][0.14\linewidth]{0.31\linewidth}
      \centering{Axial 2-chamber view}
      \\
      \smallskip
      \begin{minipage}[b]{0.32\linewidth}
      \includegraphics[width=1\linewidth]{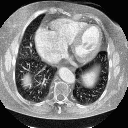} \\
      \centering{Image}
      \end{minipage}
      \begin{minipage}[b]{0.32\linewidth}
      \includegraphics[width=1\linewidth]{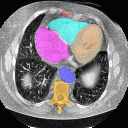}
      \centering{Ground}
      \end{minipage}
      \begin{minipage}[b]{0.32\linewidth}
      \includegraphics[width=1\linewidth]{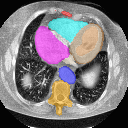}
      \centering{Seg}
      \end{minipage}
    \end{minipage}\noindent
    }
    \caption{Semantic cardiac CTA image segmentation on unseen samples with the semantic axial level shown on top of each sub-figure. The segmentations were performed on image intensity in Hounsfield unit, while the images shown in this figure were contrast-enhanced by CLAHE for visualization.}
    \label{fig:cardiac:seg}
\end{figure*}

\begin{figure*}[t]
    \scriptsize
    \centering
    \fbox{\noindent
    \fontsize{6}{7}\selectfont
    \begin{minipage}[b][0.14\linewidth]{0.31\linewidth}
      \centering{Thoracic inlet}
      \\
      \smallskip
      \begin{minipage}[b]{0.32\linewidth}
      \includegraphics[width=1\linewidth]{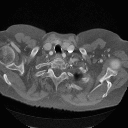} \\
      \centering{Image}
      \end{minipage}
      \begin{minipage}[b]{0.32\linewidth}
      \includegraphics[width=1\linewidth]{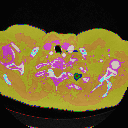}
      \centering{Ground}
      \end{minipage}
      \begin{minipage}[b]{0.32\linewidth}
      \includegraphics[width=1\linewidth]{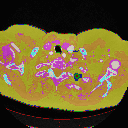}
      \centering{Seg}
      \end{minipage}
    \end{minipage}\noindent
    }
    \fbox{\noindent
    \fontsize{6}{7}\selectfont
    \begin{minipage}[b][0.14\linewidth]{0.31\linewidth}
      \centering{Lung apex}
      \\
      \smallskip
      \begin{minipage}[b]{0.32\linewidth}
      \includegraphics[width=1\linewidth]{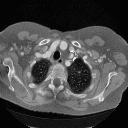} \\
      \centering{Image}
      \end{minipage}
      \begin{minipage}[b]{0.32\linewidth}
      \includegraphics[width=1\linewidth]{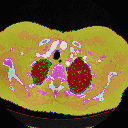}
      \centering{Ground}
      \end{minipage}
      \begin{minipage}[b]{0.32\linewidth}
      \includegraphics[width=1\linewidth]{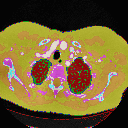}
      \centering{Seg}
      \end{minipage}
    \end{minipage}\noindent
    }
    \fbox{\noindent
    \fontsize{6}{7}\selectfont
    \begin{minipage}[b][0.14\linewidth]{0.31\linewidth}
      \centering{Great vessels origin}
      \\
      \smallskip
      \begin{minipage}[b]{0.32\linewidth}
      \includegraphics[width=1\linewidth]{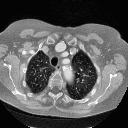} \\
      \centering{Image}
      \end{minipage}
      \begin{minipage}[b]{0.32\linewidth}
      \includegraphics[width=1\linewidth]{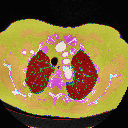}
      \centering{Ground}
      \end{minipage}
      \begin{minipage}[b]{0.32\linewidth}
      \includegraphics[width=1\linewidth]{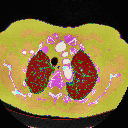}
      \centering{Seg}
      \end{minipage}
    \end{minipage}\noindent
    }
    \\
    \centering
    \fbox{\noindent
    \fontsize{6}{7}\selectfont
    \begin{minipage}[b][0.14\linewidth]{0.31\linewidth}
      \centering{Aortic arch}
      \\
      \smallskip
      \begin{minipage}[b]{0.32\linewidth}
      \includegraphics[width=1\linewidth]{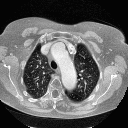} \\
      \centering{Image}
      \end{minipage}
      \begin{minipage}[b]{0.32\linewidth}
      \includegraphics[width=1\linewidth]{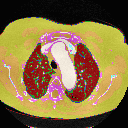}
      \centering{Ground}
      \end{minipage}
      \begin{minipage}[b]{0.32\linewidth}
      \includegraphics[width=1\linewidth]{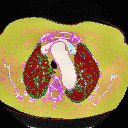}
      \centering{Seg}
      \end{minipage}
    \end{minipage}\noindent
    }
    \fbox{\noindent
    \fontsize{6}{7}\selectfont
    \begin{minipage}[b][0.14\linewidth]{0.31\linewidth}
      \centering{Ascending/descending aorta}
      \\
      \smallskip
      \begin{minipage}[b]{0.32\linewidth}
      \includegraphics[width=1\linewidth]{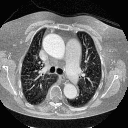} \\
      \centering{Image}
      \end{minipage}
      \begin{minipage}[b]{0.32\linewidth}
      \includegraphics[width=1\linewidth]{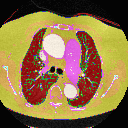}
      \centering{Ground}
      \end{minipage}
      \begin{minipage}[b]{0.32\linewidth}
      \includegraphics[width=1\linewidth]{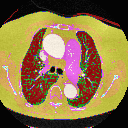}
      \centering{Seg}
      \end{minipage}
    \end{minipage}\noindent
    }
    \fbox{\noindent
    \fontsize{6}{7}\selectfont
    \begin{minipage}[b][0.14\linewidth]{0.31\linewidth}
      \centering{Pulmonary trunk}
      \\
      \smallskip
      \begin{minipage}[b]{0.32\linewidth}
      \includegraphics[width=1\linewidth]{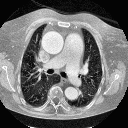} \\
      \centering{Image}
      \end{minipage}
      \begin{minipage}[b]{0.32\linewidth}
      \includegraphics[width=1\linewidth]{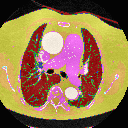}
      \centering{Ground}
      \end{minipage}
      \begin{minipage}[b]{0.32\linewidth}
      \includegraphics[width=1\linewidth]{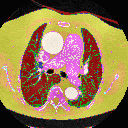}
      \centering{Seg}
      \end{minipage}
    \end{minipage}\noindent
    }
    \\
    \centering
    \fbox{\noindent
    \fontsize{6}{7}\selectfont
    \begin{minipage}[b][0.14\linewidth]{0.31\linewidth}
      \centering{Aortic valve/root}
      \\
      \smallskip
      \begin{minipage}[b]{0.32\linewidth}
      \includegraphics[width=1\linewidth]{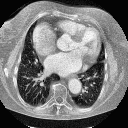} \\
      \centering{Image}
      \end{minipage}
      \begin{minipage}[b]{0.32\linewidth}
      \includegraphics[width=1\linewidth]{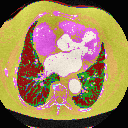}
      \centering{Ground}
      \end{minipage}
      \begin{minipage}[b]{0.32\linewidth}
      \includegraphics[width=1\linewidth]{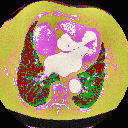}
      \centering{Seg}
      \end{minipage}
    \end{minipage}\noindent
    }
    \fbox{\noindent
    \fontsize{6}{7}\selectfont
    \begin{minipage}[b][0.14\linewidth]{0.31\linewidth}
      \centering{Axial 4-chamber view}
      \\
      \smallskip
      \begin{minipage}[b]{0.32\linewidth}
      \includegraphics[width=1\linewidth]{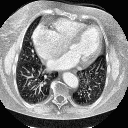} \\
      \centering{Image}
      \end{minipage}
      \begin{minipage}[b]{0.32\linewidth}
      \includegraphics[width=1\linewidth]{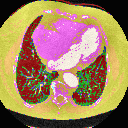}
      \centering{Ground}
      \end{minipage}
      \begin{minipage}[b]{0.32\linewidth}
      \includegraphics[width=1\linewidth]{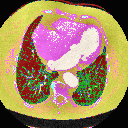}
      \centering{Seg}
      \end{minipage}
    \end{minipage}\noindent
    }
    \fbox{\noindent
    \fontsize{6}{7}\selectfont
    \begin{minipage}[b][0.14\linewidth]{0.31\linewidth}
      \centering{Axial 2-chamber view}
      \\
      \smallskip
      \begin{minipage}[b]{0.32\linewidth}
      \includegraphics[width=1\linewidth]{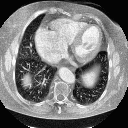} \\
      \centering{Image}
      \end{minipage}
      \begin{minipage}[b]{0.32\linewidth}
      \includegraphics[width=1\linewidth]{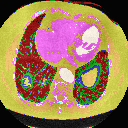}
      \centering{Ground}
      \end{minipage}
      \begin{minipage}[b]{0.32\linewidth}
      \includegraphics[width=1\linewidth]{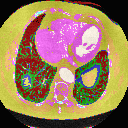}
      \centering{Seg}
      \end{minipage}
    \end{minipage}\noindent
    }
    \caption{Non-semantic cardiac CTA image segmentation on unseen samples with the semantic axial level shown on top of each sub-figure. The segmentations were performed on image intensity in Hounsfield unit, while the images shown in this figure were contrast-enhanced by CLAHE for visualization.}
    \label{fig:cardiac:seg_nonsemantic}
\end{figure*}

\begin{figure*}[p]
    \scriptsize
    \centering
    \begin{minipage}[b]{0.32\linewidth}
      \centering
      \begin{minipage}[b]{\linewidth}
      \includegraphics[width=1\linewidth]{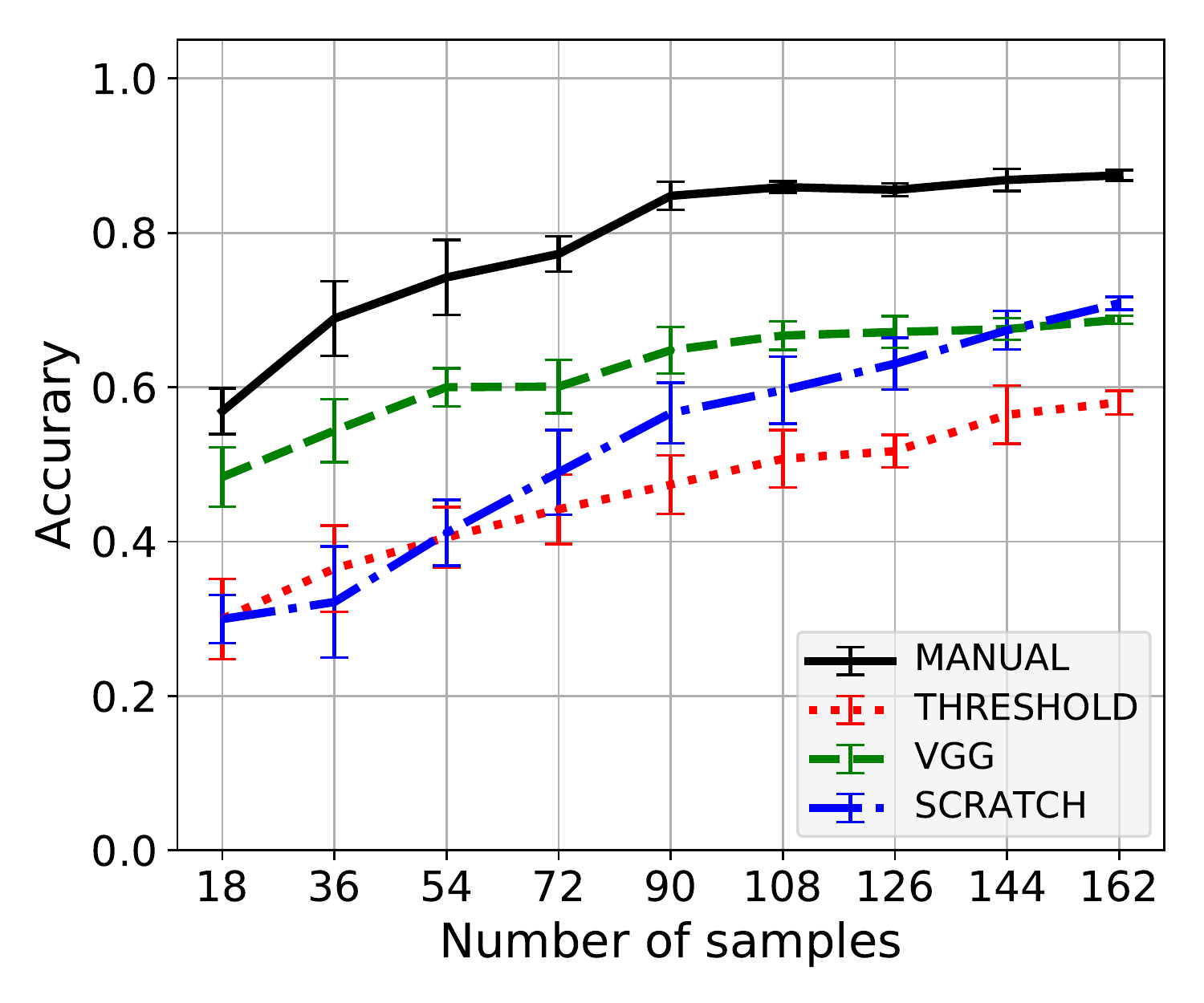}
      \centering{(a) Accuracy.}
      \end{minipage}
    \end{minipage}
    \begin{minipage}[b]{0.32\linewidth}
      \centering
      \begin{minipage}[b]{\linewidth}
      \includegraphics[width=1\linewidth]{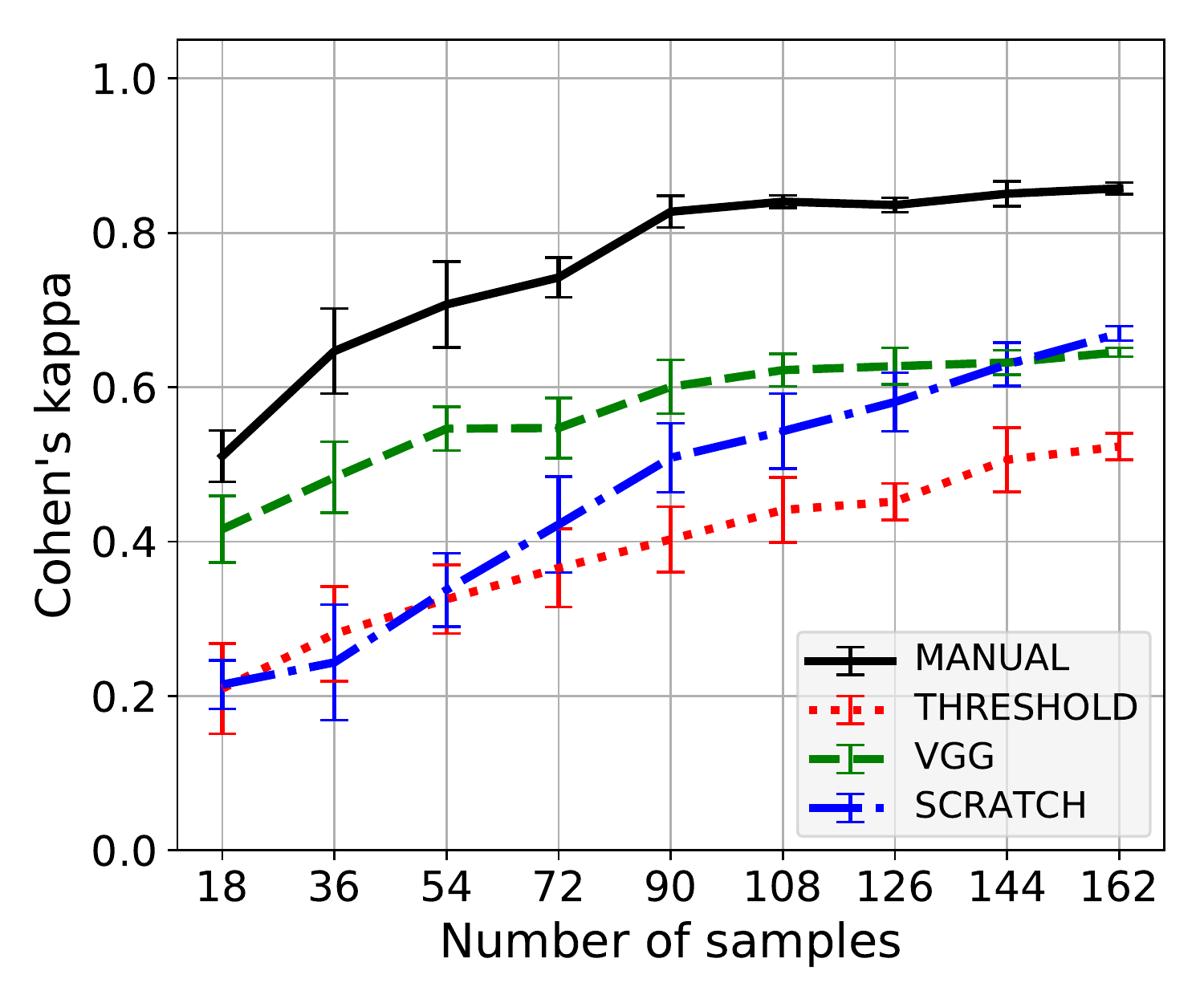}
      \centering{(b) Cohen's kappa.}
      \end{minipage}
    \end{minipage}
    \\
    \centering
    \begin{minipage}[b]{\linewidth}
      \includegraphics[width=1\linewidth]{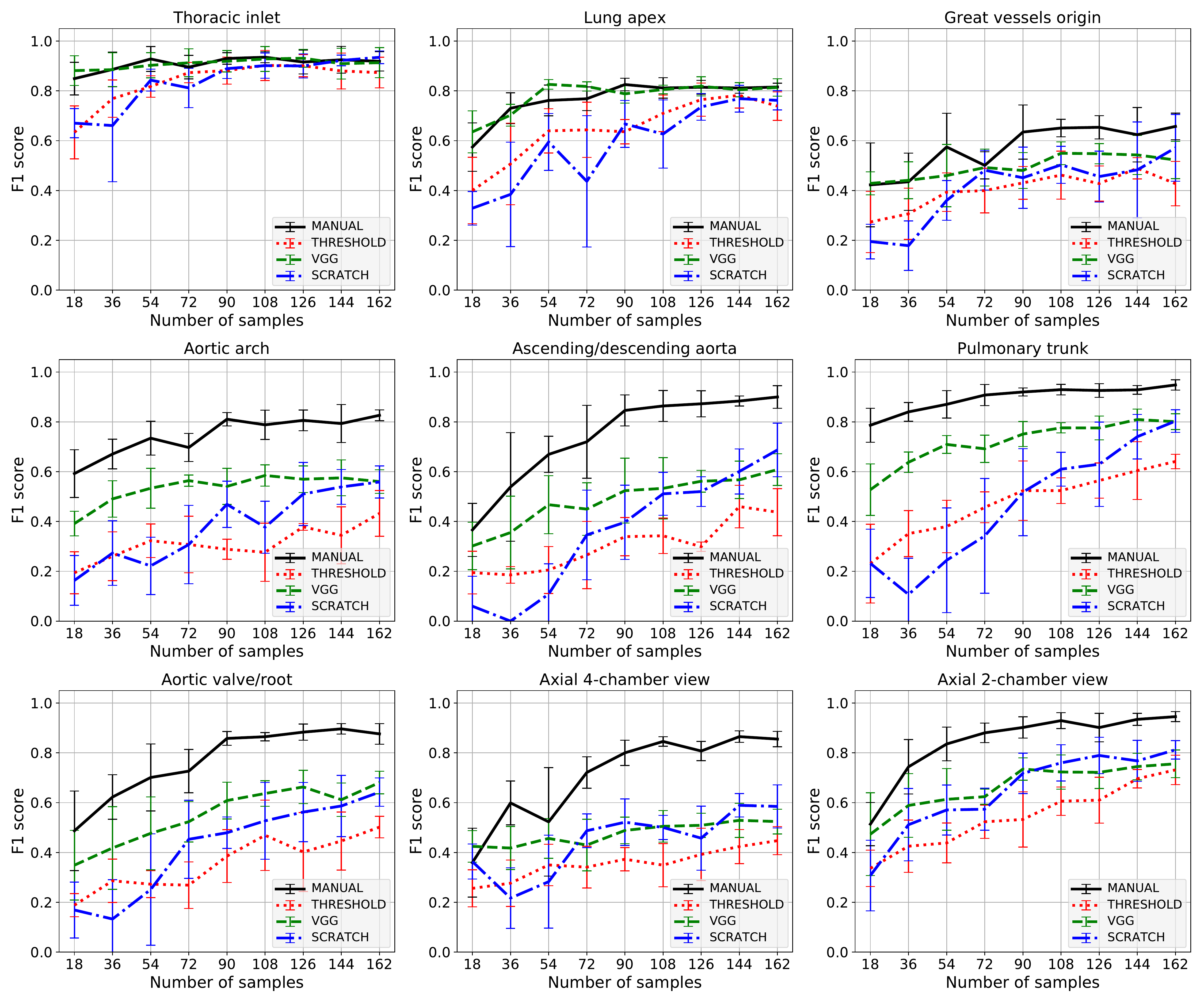}
      \centering{(c) F1 score.}
    \end{minipage}
    \caption{Cardiac semantic axial level classification. Results of the tested frameworks (Section \ref{sec:frameworks}). Each data point corresponds to five repeated experiments, with the lines representing the average values and the error bars representing the standard deviations. The $x$-axis represents the number of training samples (two to 18 for each class). The accuracies and Cohen's kappas were computed from all testing samples, while the F1 scores were computed for each class.}
    \label{fig:cardiac_classification}
\end{figure*}

\section{Semantic axial level classification on cardiac CTA images}
\label{sec:cardiac}

Apart from disease classification, we also tested our frameworks on semantic axial level classification on cardiac CTA images. With the advancements of medical imaging, 3D cardiac CTA images with hundreds of slices are common. Although these images provide valuable clinical information, they also pose challenges on computational time and accuracy for computer-aided diagnosis, especially on 3D analysis. As a result, it is beneficial to divide the entire volume into subregions. Following the work in \cite{Conference:Moradi:ISBI2016}, we tested the frameworks on classifying 2D cardiac CTA slices into nine semantic axial levels (Fig. \ref{fig:cardiac:seg}). This is a challenging task as seven of the levels are over the relatively narrow span of the heart.

\subsection{Segmentation network pre-training}

\subsubsection{Data and training}
\label{sec:cardiac_data_training}

A dataset of 48 3D cardiac CTA images was manually labeled by a radiologist to provide the training and testing data for the segmentation network. There were 16 anatomical labels in each volume (Table \ref{table:cardiac:semantic_dice}).

The volumes were split in half for training and testing, resulting in 3910 and 5215 2D slices for training and testing, respectively. All slices were resized to 128$\times$128 pixels, and we directly used the image intensity in Hounsfield unit. The segmentation network was trained with a batch size of 5, learning rate of 10$^{-3}$ and 20 epochs. Image augmentation was used. We used $n_s = 16$ and $N_s = 17$ with the background included (Fig. \ref{fig:MNet}). These hyperparameters were chosen as they provided the best performance among the tested parameters.

\subsubsection{Semantic segmentation results}

As segmentation performance is not the main concern of this paper, only the results of the trained segmentation network that was used by the classification network are presented. Table \ref{table:cardiac:semantic_dice} shows the Dice coefficients between the prediction and ground truth on the 5215 testing slices. As not all semantic labels were available on every slice, we used the pixel labels of all slices together when computing the Dice coefficients to avoid ambiguity.

There were 13 labels with Dice coefficients $\geq$74\% and the rests were $\geq$62\%. The average Dice coefficient was 78\%. The atria and the myocardium had the best Dice coefficients, followed by the pulmonary arteries, ventricles, and aorta. The sternum, aortic root, and inferior vena cava were challenging for the segmentation network. As the network could provide segmentations similar to those of the ground truth given the complicated structures across different levels of the cardiac CTA images, it contained useful features for the sematic axial level classification.

Fig. \ref{fig:cardiac:seg} shows the segmentation results on unseen samples. Although the segmentation network could not provide segmentations as detailed as the ground truth, the segmented structures were consistent with those of the ground truth with good overlaps. For the thoracic inlet, the segmentation network correctly segmented the vertebrae even it did not exist in the ground truth. These show that the network learned the structural features.

\subsubsection{Non-semantic segmentation results}

A non-semantic segmentation network was also trained for the THRESHOLD framework (Section \ref{sec:frameworks}). The same images and training strategy in Section \ref{sec:cardiac_data_training} were used to train this segmentation network, with the ground-truth labels provided by the $k$-means method with ten classes. The number of classes was chosen so that some segmented structures such as the blood pools and myocardium were visually distinguishable. Nevertheless, this was not an easy task as different structures share similar intensity values. The segmentation results on unseen samples are shown in Fig. \ref{fig:cardiac:seg_nonsemantic}. Only a few structures such as the blood pools were visible and most classes were semantically irrelevant.

\subsection{Cardiac semantic axial level classification}

\subsubsection{Data and training}

There were 425 2D slices annotated by a radiologist into nine classes of semantic axial levels for training and testing. Using 18 training samples for each class, we had 162 and 263 images for training and testing, respectively.

Similar to the segmentation network, all images were resized to 128$\times$128 pixels, and the image intensity in Hounsfield unit was used except for the VGG framework (Section \ref{sec:training_strategy}). Image augmentation was used to learn invariant features and reduce overfitting. The classification networks were trained with the batch size of 64, learning rate of 5$\times$10$^{-4}$, and 300 epochs. We used $r = 3$ for the MANUAL and THRESHOLD frameworks (Fig. \ref{fig:ClassNets}(a)), and used $n_c = 16$ and $n_{FC} = 100$ for all frameworks (Fig. \ref{fig:ClassNets}). $N_c = 9$ for the nine classes. These hyperparameters were chosen as they provided the best performance among the tested parameters. To study the performance with limited numbers of samples, we gradually increased the training samples of each class from two to 18 with strides of two (i.e. from 18 to 162 with strides of 18 for all classes). Five repetitions were performed on randomized splits of training and testing data. Identical data and setting were applied to all frameworks.

\subsubsection{Classification results}

Fig. \ref{fig:cardiac_classification} shows the results of the tested frameworks. The performances were assessed by the classification accuracy, Cohen's kappa, and F1 score. For each framework, the accuracies and Cohen's kappas were computed for the overall performance, while the F1 scores were computed for each class.

Both accuracies and Cohen's kappas show that the MANUAL framework had the best performance. The VGG framework outperformed the THRESHOLD and SCRATCH frameworks with small training samples, but was catched up by the SCRATCH framework at 144 samples. Both MANUAL and VGG frameworks converged at about 108 samples, though the MANUAL framework had better performance (86\% accuracy, 84\% Cohen's kappa) than the VGG framework (67\% accuracy, 62\% Cohen's kappa).

The F1 scores at different semantic axial levels provide further comparisons. The MANUAL framework outperformed the other frameworks in general, but the degrees were different among classes. For example, all frameworks performed best at the thoracic inlet level and achieved almost 90\% of F1 scores. A similar trend can be observed at the lung apex level. On the other hand, the differences between the MANUAL and the other frameworks were large at the ascending/decending aorta level and the 4-chamber view level. At most levels, the SCRATCH framework started with relatively low F1 scores, but gradually outperformed the THRESHOLD framework and catched up with the VGG framework with increased training samples. At 90 samples, the MANUAL framework had high average F1 scores between 80\% and 93\% except at the great vessels origin level (63\%). The MANUAL framework was more consistent than the other frameworks given its small standard deviations at all levels after convergence at 90 samples.

\section{Discussion}
\label{sec:discussion}

The experimental results in Section \ref{sec:brain} and \ref{sec:cardiac} provide several interesting observations for discussion.

As segmentation accuracy is not our main concern, we simply modified an existing network architecture (M-Net) as a feature source with respect to the computational feasibility. Although we examined different hyperparameters and selected those provided the best results, our segmentation accuracy could not be as good as those tailored for specific problems \citep{Conference:Mehta:ISBI2017,Conference:Roy:MICCAI2017}. Nevertheless, given the complicated anatomical structures with more than 16 labels, the average Dice coefficients of 69\% on 3D brain MR images and 78\% on 2D cardiac CTA images show that the segmentation networks learned useful features that improved classification performance with limited data.

From our experience on the presented segmentation and classification problems on medical images, we may not need large networks and big data to provide useful clinical applications. Using the 2D cardiac semantic axial level classification as an example, compared with the $\sim$138 million parameters of VGGNet, there were only $\sim$3 million, $\sim$17 million, and $\sim$1 million parameters for the MANUAL, VGG, and SCRATCH frameworks, respectively. This is mainly due to the fact that we usually do not have problems with 1000 classes in medical image analysis. Although we may not require a lot of training data, a reasonably large amount of testing data should be used to ensure the practical applicability of the trained models.

The MANUAL and THRESHOLD frameworks can be very useful and important especially for 3D image classification, as 3D models pre-trained from hundreds of thousands of 3D images are not publicly available to the best of our knowledge. Furthermore, notice that for the brain segmentation networks, although only trained on 21 volumes of normal cases, the learned features improved the performance on tumor type classification. This shows that instead of using hundreds of thousands of image-based annotations to provide pre-trained feature sources, we can use fewer images by using more complex pixel-based annotations. This also shows that segmentation networks trained on normal images can improve disease classification.

The results of the brain and heart classification in Fig. \ref{fig:brain:classification} and \ref{fig:cardiac_classification} show some fluctuations of the curves when the numbers of samples were small, especially for the SCRATCH framework. One of the reasons was overfitting as there were not enough data for the networks to learn probably. As the SCRATCH framework does not have pre-trained features, it had larger fluctuations and standard deviations in general. On the other hand, frameworks with pre-trained features had relatively smooth curves. This shows that using pre-trained features can reduce overfitting.

The experiments on the THRESHOLD framework were used to investigate alternatives to the time consuming and expensive manual segmentation. This framework had different performances on different datasets, and this is probably related to the natures of the problems. For the brain tumor classification, detailed brain structures may be unnecessary as long as the tumor can be distinguished, and Fig. \ref{fig:brain:seg_nonsemantic} shows that this purpose was achieved by image thresholding. On the other hand, the semantic level classification required more structural details, and image thresholding could not provide them. As potential future work, instead of using image thresholding, similar to \cite{Conference:Roy:MICCAI2017}, we can use labels generated by an existing automated software to provide good enough training data to avoid manual segmentation.

The VGG framework tested in this paper may not be optimal for the presented classification problem. In \cite{Conference:Yosinski:NIPS2014}, it was shown that features of deeper layers may be too specific to be used by another problem. As it is nontrivial to know which layers should be used and the main goal of this paper is not to explore the best way of using pre-trained features from ImageNet, we used the setting that is commonly used with limited data.

Given the large number of segmentation and classification deep learning architectures available, we could only choose the well-known architectures such as the VGGNet in this study. As different hyperparameters may lead to different results, for each framework on each dataset, we performed experiments to search for the appropriate parameters. Using the 2D cardiac semantic axial level classification as an example, the original VGGNet architecture with $n_c = 64$ and $n_{FC} = 4096$ (Fig. \ref{fig:ClassNets}(b)) was trained from scratch, and it gave very low accuracy as expected as there were not enough data to train the large number of parameters. After trying different parameters, we found that $n_c = 16$ and $n_{FC} = 100$ gave the best results and thus we report this setting in the paper. A similar procedure was performed on the 3D brain tumor type classification to get $n_c = 16$ and $n_{FC} = 200$. As there are still no computationally feasible algorithms for hyperparameter optimization to the best of our knowledge, there is no guarantee that the hyperparameters are optimal, and this is a common problem for deep learning related studies. On the other hand, we tried our best to ensure meaningful comparisons. For example, the MANUAL and THRESHOLD frameworks were identical in architecture, hyperparameters, and training, with the only difference as the ground-truth annotations for training the segmentation networks.

Although the experimental results were promising, the proposed end-to-end framework may have reduced performance when applying to image classification tasks with subtle morphological differences. In such situations, algorithms that use localized CNN features such as that in \cite{Journal:Li:SR2017} can be good alternatives.

\section{Conclusion}
\label{sec:conclusion}

In this paper, we propose a strategy for building medical image classifiers from pre-trained segmentation networks. Using a segmentation network pre-trained on data similar to those of the classification task, we can build classifiers that achieve high performance with very limited data. The proposed framework improved accuracy on 3D brain tumor classification from MR images, and outperformed the framework with ImageNet pre-trained VGGNet on 2D cardiac semantic level classification from CTA images. As this framework is applicable to 3D medical image analysis in which models equivalent to ImageNet pre-trained CNNs are unavailable, this framework provides a useful and practical alternative.

\section*{References}

\bibliography{Ref}

\end{document}